\documentclass[11pt]{article}

\usepackage[final]{acl}

\usepackage{times}
\usepackage{latexsym}
\usepackage{microtype}
\usepackage{inconsolata}

\usepackage{amsmath}
\usepackage{amssymb}
\usepackage{mathtools}

\usepackage{booktabs}
\usepackage{multirow}
\usepackage{array}
\usepackage{tabularx}
\usepackage{enumitem}

\usepackage{graphicx}

\graphicspath{
  {typepredictor_figures/}
}

\usepackage{tikz}
\usepackage{pgfplots}

\pgfplotsset{compat=1.18}

\usetikzlibrary{
  arrows.meta,
  positioning,
  shapes.geometric,
  shapes.multipart,
  fit,
  calc,
  backgrounds
}

\usepackage{fontspec}
\usepackage{polyglossia}

\setdefaultlanguage{english}
\setotherlanguage{arabic}

\newfontfamily\arabicfont[
  Script=Arabic,
  Scale=MatchLowercase,
  BoldFont=Amiri-Bold.ttf,
  ItalicFont=Amiri-Italic.ttf,
  BoldItalicFont=Amiri-BoldItalic.ttf
]{Amiri-Regular.ttf}

\newfontfamily\arabicfonttt[
  Script=Arabic,
  Scale=MatchLowercase,
  BoldFont=Amiri-Bold.ttf,
  ItalicFont=Amiri-Italic.ttf,
  BoldItalicFont=Amiri-BoldItalic.ttf
]{Amiri-Regular.ttf}

\usepackage{xurl}

\newcommand{\norel}{\textit{no\_relation}}
\newcommand{\rel}[1]{\texttt{#1}}

\newcommand{\wojoodrelations}{\ensuremath{\mathrm{Wojood}^{\mathrm{Relations}}}}

\title{U4RASD at KnowledgeGraphEval-2026:
Ontology-Constrained Encoder and LLM Systems for Arabic Relation Extraction}

\author{
\textbf{Ahmad Abbas$^{1}$, Hussein Mazeh$^{1}$, Chadi Abou Chakra$^{1}$,} \\
\textbf{Aya Jouni$^{1}$, Mohammad Ballout$^{2}$, Fadi A. Zaraket$^{1}$} \\[3pt]
$^{1}$Arab Center for Research and Policy Studies, Doha \\
$^{2}$American University of Beirut, Beirut, Lebanon \\[3pt]
\texttt{\{aabbas,hmazeh,cabouchakr,ajouni,fzaraket\}@dohainstitute.edu.qa} \\
\texttt{mb185@aub.edu.lb}
}

\begin{document}

\maketitle

\begin{abstract}
This paper details our submission to KnowledgeGraphEval 2026 Subtask 2 on directed Arabic relation extraction. We explored encoder-based and large language model approaches for predicting one of 40 relations or \norel{} from an Arabic sentence and a subject–object pair. Our encoder systems use typed entity markers and separate relation-existence and relation-classification heads, while our LLM systems fine-tune Yehia and Qwen3.5 with QLoRA and ontology-constrained prediction. Our best system, an ensemble of Yehia-RE, Qwen-RE, and TypedPair, achieved a micro-F1 score of 94.06 and ranked \textbf{1st} in Subtask~2. Yehia-RE (full-vocabulary) achieved 92.98, while the best encoder and
Qwen-RE achieved 92.69 and 91.94 respectively. Our results highlight the importance of constrained prediction and complementary modeling approaches for Arabic relation extraction.
\end{abstract}

\section{Introduction}
\label{sec:introduction}

Relation Extraction (RE) is an information extraction task that aims to identify semantic relations between entities mentioned in text. In directed RE, the relation is predicted from a designated subject to a designated object. Arabic RE remains challenging when relations are semantically similar, entity mentions occur multiple times in the same sentence, or entity-type information is needed to determine which relations are valid.

In this paper, we present our submission to KnowledgeGraphEval 2026 Subtask~2 on directed Arabic relation extraction. Given an Arabic sentence and a subject--object pair, the task is to predict one of 40 positive relations or \norel{} when no valid relation is expressed \citep{knowledgegrapheval2026}. The task data are also highly imbalanced, making accurate positive-relation prediction and \norel{} handling particularly important.

We investigate two system families. The first, TypedPair, uses Arabic encoder models with typed directional entity markers and separate heads for relation existence and positive-relation classification. The second uses Yehia \citep{yehia2025} and Qwen3.5 \citep{qwen35}, adapted with QLoRA \citep{dettmers2023qlora}, resulting in our Yehia-RE and Qwen-RE systems. Yehia-RE formulates relation prediction as a constrained Arabic one-token multiple-choice task, while Qwen-RE is evaluated using both causal-LM prediction and sequence classification. All systems share entity-occurrence resolution, entity-type prediction, and ontology-based relation filtering.

Our main contributions are: (1) the TypedPair encoder family; (2)
ontology-constrained Yehia-RE and Qwen-RE systems; and (3) a shared pipeline
for occurrence resolution, entity typing, and relation filtering. Our best
ensemble achieved a micro-$F_1$ score of 94.06 and ranked \textbf{1st} in
Subtask~2. The strongest individual Yehia-RE, TypedPair, and Qwen-RE systems
achieved 92.98, 92.69, and 91.94, respectively. The trained models and
released resources are available at \url{https://huggingface.co/U4RASD}.
\section{Background}
\label{sec:background}

The task data are derived from \wojoodrelations{}
\citep{aljabari-etal-2025-wojoodrelations}, while WojoodOntology provides
relation definitions and domain--range constraints over entity types
\citep{aljabari-etal-2025-wojoodontology}. Konooz
\citep{hamad-etal-2025-konooz} is used in the shared task's NER subtask;
our submission addresses only Subtask~2 on relation extraction.

Prior relation-extraction work has explored marker-based encoder models, including entity markers \citep{baldini-soares-etal-2019-matching}, PL-Marker \citep{ye-etal-2022-packed}, and typed entity markers \citep{zhou-chen-2022-improved}. For Arabic RE, \wojoodrelations{} introduces NLI-RE, which formulates candidate relations as natural-language-inference hypotheses and filters them according to compatible entity types \citep{aljabari-etal-2025-wojoodrelations}.

Generative approaches have also modeled RE through sequence generation and
prompting. AREEj augments Arabic relation extraction with textual evidence supervision \citep{mraikhat-etal-2024-areej}, while GPT-RE and related work explore in-context and instruction-based relation prediction \citep{wan-etal-2023-gpt,wadhwa-etal-2023-revisiting,
zhang-etal-2023-aligning}. Our work investigates both encoder-based and constrained LLM-based approaches under the same ontology-aware Arabic RE setting.
\vspace{-2.0mm}
\section{System Overview}
\label{sec:system-overview}
\vspace{-2.0mm}

\begin{figure*}[t]
\centering
\begin{tikzpicture}[
  node distance=0.65cm and 0.65cm,
  stage/.style={
    draw=gray!70,
    fill=gray!5,
    rounded corners=2pt,
    align=center,
    minimum height=0.82cm,
    text width=2.35cm,
    font=\small
  },
  family/.style={
    draw=blue!55!black,
    fill=blue!5,
    rounded corners=3pt,
    align=center,
    minimum height=0.95cm,
    text width=3.15cm,
    font=\small
  },
  result/.style={
    draw=black,
    fill=green!8,
    rounded corners=3pt,
    align=center,
    minimum height=0.82cm,
    text width=2.55cm, 
    font=\small\bfseries
  },
  arrow/.style={
    -{Latex[length=2mm,width=1.4mm]},
    line width=0.6pt,
    draw=gray!80!black
  }
]

\node[stage] (input)
  {Sentence and directed\\subject--object pair};

\node[stage, right=of input] (spans)
  {Exact occurrence\\resolution};

\node[stage, right=of spans] (types)
  {21-way entity-type\\prediction};

\node[stage, right=of types] (ontology)
  {Ontology-compatible\\relation set};

\node[family, below left=1.0cm and 0.35cm of ontology] (encoder)
  {\textbf{Encoder family}\\TypedPair two-head models};

\node[family, below right=1.0cm and 0.35cm of ontology] (llm)
  {\textbf{LLM family}\\Yehia-RE and Qwen-RE};

\node[result, below=2.65cm of ontology] (output)
  {Final relation};

\draw[arrow] (input) -- (spans);
\draw[arrow] (spans) -- (types);
\draw[arrow] (types) -- (ontology);

\draw[arrow] (ontology.south) -- ++(0,-0.35)
  -| (encoder.north);

\draw[arrow] (ontology.south) -- ++(0,-0.35)
  -| (llm.north);

\draw[arrow] (encoder.south) -- ++(0,-0.35)
  |- (output.west);

\draw[arrow] (llm.south) -- ++(0,-0.35)
  |- (output.east);

\end{tikzpicture}
\caption{Shared relation-extraction pipeline and the two model families.}
\label{fig:system-overview}
\end{figure*}
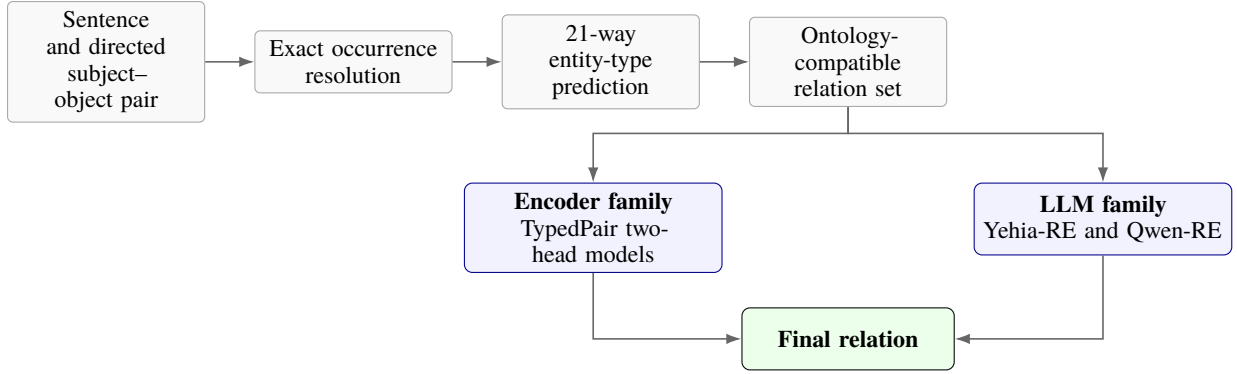

Figure~\ref{fig:system-overview} summarizes the shared full pipeline and the two model families. The following subsections summarize the components that distinguish the submitted systems.


\subsection{Shared Pipeline}
\label{sec:shared-pipeline}

The input provides the subject and object strings, but repeated occurrences may require resolving which mentions should be evaluated, and entity types are not directly provided to the relation models. We therefore locate all exact subject and object occurrences and construct the possible directed occurrence pairs. Each occurrence is then assigned an entity type using our TypePredictor, and the predicted types are used both in the typed entity markers and to restrict the candidate relations according to WojoodOntology. When multiple occurrence pairs are possible, the system applies a row-level
decision strategy to select or combine their predictions. TypePredictor and
candidate-pair handling are detailed in Appendices~\ref{app:typepredictor}
and~\ref{app:span-resolution}.
 
\subsection{Encoder-Based Systems}
\label{sec:encoder-main}

The TypedPair family inserts typed directional markers around the selected subject and object occurrences. It combines the sentence-start state, the two opening-marker states, simple pair interactions, and a 42-dimensional subject--object type vector. A multilayer perceptron feeds two heads: a binary head predicts whether a positive relation exists, and a 40-way head selects the positive relation. Negative rows supervise only the existence head, avoiding an arbitrary positive-class target. At inference, the positive head is masked by the predicted type pair and is consulted only when the existence score passes a validation-selected threshold.

We evaluate NeoAraBERT \citep{abou-chakra-etal-2026-neoarabert},
AraBERTv02 \citep{antoun-etal-2020-arabert},
ARBERTv2 \citep{elmadany-etal-2023-orca},
AraBERTv02-Large \citep{antoun-etal-2020-arabert},
and AraModernBERT \citep{elshehy-etal-2026-aramodernbert} backbones, together with various row-level candidate-pair decision strategies. The exact representation, loss, threshold selection, tensor dimensions, and candidate algorithms are provided in Appendix~\ref{app:TypedPair}.

\subsection{LLM-Based Systems}
\label{sec:llm-main}

\subsubsection{Yehia-RE}

We fine-tune Yehia with QLoRA and formulate each relation decision as a
row-specific Arabic multiple-choice problem. The marked sentence, predicted
entity types, and only the ontology-compatible positive relations are shown in
the prompt, followed by \norel{}. Each displayed option is mapped to an
existing Arabic one-token code, and the model returns the code alone.

We evaluate two Yehia-RE training variants. The full-vocabulary variant optimizes the gold code against the complete model vocabulary and, at inference, compares only the displayed option codes. The row-local variant instead normalizes the training objective only over the option codes displayed for the current example. Full objectives and candidate-pair aggregation are detailed in
Appendix~\ref{app:yehia}.

\subsubsection{Qwen-RE}

We fine-tune Qwen3.5-4B and Qwen3.5-9B with QLoRA using prompts containing the marked Arabic sentence, entity strings and types, and concise meanings for the ontology-compatible labels. We denote the causal-LM variants as Qwen-RE-4B and Qwen-RE-9B, and the 9B sequence-classification variant as Qwen-RE-9B-Cls. The markers identify the entity occurrences in the sentence and make the subject--object direction explicit. The causal-LM systems are trained to produce the relation label. At inference, we either generate the label or append each compatible label and score it using its mean token log-probability. Because the output label set is fixed, we additionally fine-tune Qwen3.5-9B with a trainable 41-way sequence-classification head. This classifier receives a shorter prompt and directly predicts a relation class, with ontology-incompatible classes masked during both training and inference, avoiding textual generation and parsing failures.

The Qwen prompt structures, a paired causal-LM/classifier example, complete-label scoring, class-imbalance interventions, and training configuration appear in Appendix~\ref{app:qwen}. We also tested manual disambiguation rules and GEPA prompt evolution \citep{agrawal2026gepa}; these exploratory interventions did not replace the selected submitted configurations and are documented separately in Appendix~\ref{app:prompt-optimization}.
\section{Experimental Setup}
\label{sec:experimental-setup}

\subsection{Data and Preprocessing}

The labeled training data were enriched with occurrence-specific character
spans and entity types. We first aligned relation mentions with the Wojood flat
and nested NER annotations \citep{jarrar-etal-2022-wojood}; unresolved mentions
were then matched against predictions from the Wojood NER model. Two authors manually resolved the remaining 240 training rows that could
not be enriched automatically, retaining the original relation labels.
This enrichment was used only for training; blind-test rows were processed
entirely by the full inference pipeline. Conflict filtering, ontology compatibility filtering, and deduplication reduced 17,381 raw relation rows to 15,686 training instances; the development artifact contains 1,687 rows, while the official blind evaluation input contains 2,074 rows. The development artifact
contains both positive relations and \norel{} examples, whereas the official
blind evaluation input contains only positive-relation rows; the resulting
difference in class distribution means development and blind micro-$F_1$
scores are not directly comparable. Complete processing details are given in
Appendix~\ref{app:relation-data}.

The processed training distribution is strongly imbalanced: \norel{} has 6,677 examples, followed by \rel{Location.located\_in} with 4,269 and \rel{PartOf.geopolitical\_division} with 1,684. Encoder models use cross-entropy objectives, Yehia-RE uses frequency-based sampling, the Qwen-RE causal-LM experiments use clipped inverse-frequency weighting, and the Qwen-RE classifier uses square-root inverse-frequency sampling. Selected relation-frequency statistics are reported in Appendix~\ref{app:relation-data}, while model-specific sampling and weighting
details are reported in Appendices~\ref{app:TypedPair},
\ref{app:yehia}, and \ref{app:qwen}.

\subsection{Training, Decoding, and Selection}

A cross-system summary of the principal training and selection configurations is provided in Table~\ref{tab:main-config} in Appendix~\ref{app:reproducibility}. TypePredictor and TypedPair encoders are fully fine-tuned. Yehia and Qwen use 4-bit NF4 QLoRA with frozen base weights. Marker handling after tokenization follows the procedure described in Appendix~\ref{app:span-resolution}. Relation-extraction checkpoints are selected using the 1,687-row development artifact and then frozen before official inference. Fixed runner settings are reported separately in the corresponding system appendices. TypePredictor is selected independently using its entity-typing validation split. 
\paragraph{Efficiency.}
AraBERTv02-Large TypedPair required approximately 43 minutes on a single
NVIDIA A40, while Yehia-RE required approximately 3.32 hours on a single
NVIDIA L40S. Qwen-RE-9B required approximately 12 hours on a single NVIDIA
RTX PRO 4500. Since these measurements were obtained on different GPUs, they
should not be treated as a controlled efficiency comparison, but they
illustrate the substantially higher training cost of the LLM-based systems
relative to the encoder-based model.

Official evaluation uses micro-$F_1$. Depending on the configuration, the
row-level decision uses either majority voting, top-existence/top-confidence
selection, max-positive selection, or nearest-candidate selection.
Detailed definitions and deterministic tie-breaking are provided in
Appendices~\ref{app:span-resolution}, \ref{app:TypedPair},
\ref{app:yehia}, and \ref{app:qwen}.
\section{Results}
\label{sec:results}

Table~\ref{tab:official-results} summarizes representative official
submissions. All values are full-pipeline scores on the blind set;
post-submission prompt analyses are reported separately in
Appendix~\ref{app:prompt-optimization}. The shared task permitted up to
30 official submissions per team; we used 22.
\begin{table*}[t]
\centering
\small
\setlength{\tabcolsep}{3.5pt}
\begin{tabular}{@{}p{0.31\textwidth}p{0.28\textwidth}p{0.21\textwidth}r@{}}
\toprule
\textbf{System} & \textbf{Training/decoder} &
\textbf{Row-level decision} & \textbf{Official $F_1$} \\
\midrule
TypedPair--NeoAraBERT & Two-head encoder & Majority + logit-sum & 88.33 \\
TypedPair--ARBERTv2 & Two-head encoder & Majority + logit-sum & 91.96 \\
TypedPair--AraBERTv02-Large & Two-head encoder & Top existence & 92.29 \\
TypedPair--AraModernBERT & Two-head encoder & Majority + logit-sum & \textbf{92.69} \\
\midrule
Yehia-RE (row-local) & Row-local CE & Max positive & 92.84 \\
Yehia-RE (full-vocabulary) & Full-vocabulary CE &
Max positive & \textbf{92.98} \\
\midrule
Qwen-RE-4B & Causal-LM label generation & Nearest candidate & 91.53 \\
Qwen-RE-9B & Causal-LM label generation & Nearest candidate & 91.76 \\
Qwen-RE-9B-Cls & Sequence classification & Nearest candidate & \textbf{91.94} \\
\midrule
Team ensemble &
Yehia-RE (full-vocabulary) + Qwen-RE-9B-Cls + TypedPair--AraModernBERT &
Majority vote; Yehia-RE tie-break &
\textbf{94.06} \\
\bottomrule
\end{tabular}
\caption{Summary of representative official KnowledgeGraphEval 2026
Subtask~2 submissions. Bold values mark the strongest configuration within
each model family.}
\label{tab:official-results}
\end{table*}
\subsection{Encoder Results}

Across the official blind submissions, AraModernBERT achieved the strongest
encoder result at 92.69, compared with 91.96 for ARBERTv2 and
92.20--92.29 for the two reported AraBERTv02-Large aggregation strategies.
NeoAraBERT with majority + logit-sum obtained 88.33. A component ablation on AraModernBERT
(Appendix~\ref{app:TypedPair-ablation}) shows that the complete configuration
with both typed text markers and the directional type vector achieved the
strongest performance among the tested variants. The complete encoder and aggregation results
are reported in Appendix~\ref{app:TypedPair-results}.

\subsection{LLM Results}

Yehia-RE (full-vocabulary) achieved 92.98, while the row-local variant reached
92.84 with max-positive aggregation. Its constrained one-token formulation
eliminates free-form label formatting errors by casting relation prediction as a
row-specific Arabic multiple-choice task. Our best system, a three-system
majority ensemble of Yehia-RE, Qwen-RE, and TypedPair, achieved 94.06 and ranked
\textbf{1st} in Subtask~2. Detailed Yehia-RE formulations, candidate
aggregation, and official results are provided in Appendix~\ref{app:yehia}.

On the official blind evaluation, Qwen-RE-4B and Qwen-RE-9B obtained 91.53
and 91.76, respectively, while Qwen-RE-9B-Cls obtained 91.94. These blind-set differences are
reported descriptively rather than as controlled ablations. Complete official Qwen results are reported in Appendix~\ref{app:qwen-results}.

\subsection{Error Analysis}

The main failure modes reflect interactions among pipeline components. A type
error can change the typed markers and remove the correct relation from the
compatible set. Repeated surface forms can associate the prediction with the
wrong textual evidence. Semantically close relations, especially generic
location versus headquarters or administrative division, are difficult under a
highly skewed label distribution. Finally, causal-generation systems may understand
the relation but emit an invalid or partial canonical label. Additional qualitative error analysis is provided in
Appendix~\ref{app:error-analysis}, while the exploratory disambiguation
experiments are reported in Appendix~\ref{app:prompt-optimization}.
\section{Conclusion}
\label{sec:conclusion}

We presented two complementary system families for KnowledgeGraphEval 2026
Arabic relation extraction. The encoder family combines typed directional
markers with a two-head relation objective, while the LLM family constrains
Yehia-RE and Qwen-RE to ontology-compatible decisions. Our best system, an
ensemble of Yehia-RE, Qwen-RE, and TypedPair, achieved an official micro-$F_1$
score of 94.06 and ranked \textbf{1st} in Subtask~2. Yehia-RE
(full-vocabulary) achieved 92.98, while the strongest encoder and Qwen-RE
systems reached 92.69 and 91.94, respectively. The results highlight the
importance of constrained prediction, entity-type compatibility, candidate
resolution, and complementary modeling approaches. Future work will propagate type uncertainty, jointly score occurrence and
relation hypotheses, and study confidence-weighted ensembles.
\section*{Limitations}

The reported training-level results are based on a single random seed.
Consequently, small differences between model configurations should be
interpreted cautiously, as we cannot quantify variation across training runs.
The systems also depend on hard entity-type predictions and exact-match
occurrence recovery, and they remain challenging for rare or semantically
adjacent relations. Because the TypePredictor dataset uses a mention-level split, sentence-level
overlap with relation-extraction inputs is possible. Although TypePredictor
never uses relation labels, such overlap may affect the entity-type predictions
used for typed markers and ontology filtering.
\bibliography{custom}

@inproceedings{knowledgegrapheval2026,
  title     = {{KnowledgeGraphEval 2026}: The {F}irst {S}hared {T}ask on {A}rabic {K}nowledge {G}raphs},
  author    = {Aljabari, Alaa and
               Hamad, Nagham and
               El Mekki, Abdellah and
               Abdul-Mageed, Muhammad and
               Zitouni, Imed and
               Chawla, Sanjay and
               Jarrar, Mustafa},
  booktitle = {Proceedings of the Fourth Arabic Natural Language Processing Conference},
  year      = {2026},
  address   = {Budapest, Hungary},
  publisher = {Association for Computational Linguistics}
}

@inproceedings{aljabari-etal-2025-wojoodrelations,
  author    = {Aljabari, Alaa and Khalilia, Mohammed and Jarrar, Mustafa},
  title     = {$\mathrm{Wojood^{Relations}}$: {A}rabic Relation Extraction Corpus and Modeling},
  booktitle = {Proceedings of the 2025 Conference on Empirical Methods in Natural Language Processing},
  month     = nov,
  year      = {2025},
  address   = {Suzhou, China},
  publisher = {Association for Computational Linguistics},
  pages     = {34342--34360},
  doi       = {10.18653/v1/2025.emnlp-main.1741},
  url       = {https://aclanthology.org/2025.emnlp-main.1741/}
}

@inproceedings{aljabari-etal-2025-wojoodontology,
  author    = {Aljabari, Alaa and Hamad, Nagham and Khalilia, Mohammed and Jarrar, Mustafa},
  title     = {{W}ojood{O}ntology: Ontology-Driven {LLM} Prompting for Unified Information Extraction Tasks},
  booktitle = {Proceedings of The Third Arabic Natural Language Processing Conference},
  month     = nov,
  year      = {2025},
  address   = {Suzhou, China},
  publisher = {Association for Computational Linguistics},
  pages     = {179--193},
  doi       = {10.18653/v1/2025.arabicnlp-main.14},
  url       = {https://aclanthology.org/2025.arabicnlp-main.14/}
}

@inproceedings{jarrar-etal-2022-wojood,
  author    = {Jarrar, Mustafa and Khalilia, Mohammed and Ghanem, Sana},
  title     = {Wojood: Nested {A}rabic Named Entity Corpus and Recognition using {BERT}},
  booktitle = {Proceedings of the Thirteenth Language Resources and Evaluation Conference},
  month     = jun,
  year      = {2022},
  address   = {Marseille, France},
  publisher = {European Language Resources Association},
  pages     = {3626--3636},
  url       = {https://aclanthology.org/2022.lrec-1.387/}
}

@inproceedings{hamad-etal-2025-konooz,
  title = {Konooz: Multi-domain Multi-dialect Corpus for Named Entity Recognition},
  author = {Hamad, Nagham and
            Khalilia, Mohammed and
            Jarrar, Mustafa},
  booktitle = {Findings of the Association for Computational Linguistics: ACL 2025},
  month = jul,
  year = {2025},
  address = {Vienna, Austria},
  publisher = {Association for Computational Linguistics},
  pages = {7316--7331},
  doi = {10.18653/v1/2025.findings-acl.382},
  url = {https://aclanthology.org/2025.findings-acl.382/}
}

@inproceedings{baldini-soares-etal-2019-matching,
  author    = {Baldini Soares, Livio and FitzGerald, Nicholas and Ling, Jeffrey and Kwiatkowski, Tom},
  title     = {Matching the Blanks: Distributional Similarity for Relation Learning},
  editor = {Korhonen, Anna  and
      Traum, David  and
      M{\`a}rquez, Llu{\'i}s},
  booktitle = {Proceedings of the 57th Annual Meeting of the Association for Computational Linguistics},
  month     = jul,
  year      = {2019},
  address   = {Florence, Italy},
  publisher = {Association for Computational Linguistics},
  pages     = {2895--2905},
  doi       = {10.18653/v1/P19-1279},
  url       = {https://aclanthology.org/P19-1279/}
}

@inproceedings{ye-etal-2022-packed,
  author    = {Ye, Deming and Lin, Yankai and Li, Peng and Sun, Maosong},
  title     = {Packed Levitated Marker for Entity and Relation Extraction},
  editor = {Muresan, Smaranda  and
      Nakov, Preslav  and
      Villavicencio, Aline},
  booktitle = {Proceedings of the 60th Annual Meeting of the Association for Computational Linguistics (Volume 1: Long Papers)},
  month     = may,
  year      = {2022},
  address   = {Dublin, Ireland},
  publisher = {Association for Computational Linguistics},
  pages     = {4904--4917},
  doi       = {10.18653/v1/2022.acl-long.337},
  url       = {https://aclanthology.org/2022.acl-long.337/}
}

@inproceedings{dettmers2023qlora,
  author    = {Dettmers, Tim and Pagnoni, Artidoro and Holtzman, Ari and Zettlemoyer, Luke},
  editor = {A. Oh and T. Naumann and A. Globerson and K. Saenko and M. Hardt and S. Levine},
  title     = {{QLoRA}: Efficient Finetuning of Quantized {LLM}s},
  booktitle = {Advances in Neural Information Processing Systems},
  volume    = {36},
  pages     = {10088--10115},
  year      = {2023},
  doi       = {10.52202/075280-0441},
  url       = {https://proceedings.neurips.cc/paper_files/paper/2023/hash/1feb87871436031bdc0f2beaa62a049b-Abstract-Conference.html}
}

@inproceedings{agrawal2026gepa,
  author    = {Lakshya A Agrawal and Shangyin Tan and Dilara Soylu and Noah Ziems and Rishi Khare and Krista Opsahl-Ong and Arnav Singhvi and Herumb Shandilya and Michael J Ryan and Meng Jiang and Christopher Potts and Koushik Sen and Alex Dimakis and Ion Stoica and Dan Klein and Matei Zaharia and Omar Khattab},
  title     = {{GEPA}: Reflective Prompt Evolution Can Outperform Reinforcement Learning},
  booktitle = {The Fourteenth International Conference on Learning Representations},
  year      = {2026},
  url       = {https://openreview.net/forum?id=RQm2KQTM5r}
}

@inproceedings{zhou-chen-2022-improved,
  title     = {An Improved Baseline for Sentence-level Relation Extraction},
  author    = {Zhou, Wenxuan and Chen, Muhao},
  editor = {He, Yulan  and
      Ji, Heng  and
      Li, Sujian  and
      Liu, Yang  and
      Chang, Chua-Hui},
  booktitle = {Proceedings of the 2nd Conference of the Asia-Pacific Chapter of the Association for Computational Linguistics and the 12th International Joint Conference on Natural Language Processing (Volume 2: Short Papers)},
  month     = nov,
  year      = {2022},
  address   = {Online only},
  publisher = {Association for Computational Linguistics},
  pages     = {161--168},
  doi       = {10.18653/v1/2022.aacl-short.21},
  url       = {https://aclanthology.org/2022.aacl-short.21/}
}

@inproceedings{wan-etal-2023-gpt,
  title     = {{GPT}-{RE}: In-context Learning for Relation Extraction using Large Language Models},
  author    = {Wan, Zhen and Cheng, Fei and Mao, Zhuoyuan and Liu, Qianying and Song, Haiyue and Li, Jiwei and Kurohashi, Sadao},
  editor = {Bouamor, Houda  and
      Pino, Juan  and
      Bali, Kalika},
  booktitle = {Proceedings of the 2023 Conference on Empirical Methods in Natural Language Processing},
  month     = dec,
  year      = {2023},
  address   = {Singapore},
  publisher = {Association for Computational Linguistics},
  pages     = {3534--3547},
  doi       = {10.18653/v1/2023.emnlp-main.214},
  url       = {https://aclanthology.org/2023.emnlp-main.214/}
}

@inproceedings{zhang-etal-2023-aligning,
  title     = {Aligning Instruction Tasks Unlocks Large Language Models as Zero-Shot Relation Extractors},
  author    = {Zhang, Kai  and Jimenez Gutierrez, Bernal  and Su, Yu},
  editor = {Rogers, Anna and Boyd-Graber, Jordan  and Okazaki, Naoaki},
  booktitle = {Findings of the Association for Computational Linguistics: ACL 2023},
  month     = jul,
  year      = {2023},
  address   = {Toronto, Canada},
  publisher = {Association for Computational Linguistics},
  pages     = {794--812},
  doi       = {10.18653/v1/2023.findings-acl.50},
  url       = {https://aclanthology.org/2023.findings-acl.50/}
}

@inproceedings{wadhwa-etal-2023-revisiting,
  title     = {Revisiting Relation Extraction in the era of Large Language Models},
  author    = {Wadhwa, Somin  and Amir, Silvio  and Wallace, Byron},
  editor = {Rogers, Anna and Boyd-Graber, Jordan  and Okazaki, Naoaki},
  booktitle = {Proceedings of the 61st Annual Meeting of the Association for Computational Linguistics (Volume 1: Long Papers)},
  month     = jul,
  year      = {2023},
  address   = {Toronto, Canada},
  publisher = {Association for Computational Linguistics},
  pages     = {15566--15589},
  doi       = {10.18653/v1/2023.acl-long.868},
  url       = {https://aclanthology.org/2023.acl-long.868/}
}

@misc{yehia2025,
  title        = {Yehia 7B Preview},
  author       = {{Navid-AI}},
  year         = {2025},
  howpublished = {\url{https://huggingface.co/Navid-AI/Yehia-7B-preview}}
}

@misc{qwen35,
  title  = {{Qwen3.5}: Towards Native Multimodal Agents},
  author = {{Qwen Team}},
  month  = feb,
  year   = {2026},
  url    = {https://qwen.ai/blog?id=qwen3.5}
}

@inproceedings{mraikhat-etal-2024-areej,
  title     = {{AREE}j: {A}rabic Relation Extraction with Evidence},
  author    = {Rakan Al Mraikhat, Osama and Hamoud, Hadi and Zaraket, Fadi A.},
  editor = {Habash, Nizar  and
      Bouamor, Houda  and
      Eskander, Ramy  and
      Tomeh, Nadi  and
      Abu Farha, Ibrahim  and
      Abdelali, Ahmed  and
      Touileb, Samia  and
      Hamed, Injy  and
      Onaizan, Yaser  and
      Alhafni, Bashar  and
      Antoun, Wissam  and
      Khalifa, Salam  and
      Haddad, Hatem  and
      Zitouni, Imed  and
      AlKhamissi, Badr  and
      Almatham, Rawan  and
      Mrini, Khalil},
  booktitle = {Proceedings of the Second Arabic Natural Language Processing Conference},
  month     = aug,
  year      = {2024},
  address   = {Bangkok, Thailand},
  publisher = {Association for Computational Linguistics},
  pages     = {67--72},
  doi       = {10.18653/v1/2024.arabicnlp-1.6},
  url       = {https://aclanthology.org/2024.arabicnlp-1.6/}
}

@inproceedings{abou-chakra-etal-2026-neoarabert,
  title     = {{N}eo{A}ra{BERT}: A Modern Foundation Model for {A}rabic Embeddings with Diacritics-Aware Tokenization and {POS}-Targeted Masking},
  author    = {Abou Chakra, Chadi and Hamoud, Hadi Khaled and Rakan Al Mraikhat, Osama and Abu Obaida, Qusai and Ballout, Mohamad and Zaraket, Fadi},
  editor = {Liakata, Maria and Moreira, Viviane P.  and Zhang, Jiajun  and Jurgens, David},
  booktitle = {Findings of the {A}ssociation for {C}omputational {L}inguistics: {ACL} 2026},
  month     = jul,
  year      = {2026},
  address   = {San Diego, California, United States},
  publisher = {Association for Computational Linguistics},
  pages     = {25952--25968},
  doi       = {10.18653/v1/2026.findings-acl.1293},
  url       = {https://aclanthology.org/2026.findings-acl.1293/}
}

@inproceedings{antoun-etal-2020-arabert,
  title     = {{AraBERT}: Transformer-based Model for {A}rabic Language Understanding},
  author    = {Antoun, Wissam and Baly, Fady and Hajj, Hazem},
  editor = {Al-Khalifa, Hend and Magdy, Walid  and Darwish, Kareem and Elsayed, Tamer and Mubarak, Hamdy},
  booktitle = {Proceedings of the 4th Workshop on Open-Source Arabic Corpora and Processing Tools, with a Shared Task on Offensive Language Detection},
  month     = may,
  year      = {2020},
  address   = {Marseille, France},
  publisher = {European Language Resources Association},
  pages     = {9--15},
  url       = {https://aclanthology.org/2020.osact-1.2/}
}

@inproceedings{elmadany-etal-2023-orca,
  title     = {{ORCA}: A Challenging Benchmark for {A}rabic Language Understanding},
  author    = {Elmadany, AbdelRahim and Nagoudi, ElMoatez Billah and Abdul-Mageed, Muhammad},
  editor = {Rogers, Anna and Boyd-Graber, Jordan  and Okazaki, Naoaki},
  booktitle = {Findings of the Association for Computational Linguistics: ACL 2023},
  month     = jul,
  year      = {2023},
  address   = {Toronto, Canada},
  publisher = {Association for Computational Linguistics},
  pages     = {9559--9586},
  doi       = {10.18653/v1/2023.findings-acl.609},
  url       = {https://aclanthology.org/2023.findings-acl.609/}
}

@inproceedings{elshehy-etal-2026-aramodernbert,
  title     = {{A}ra{M}odern{BERT}: Transtokenized Initialization and Long-Context Encoder Modeling for {A}rabic},
  author    = {Elshehy, Omar and Nacar, Omer and Djamai, Abdelbasset and Ragab, Muhammed and Al Jallad, Khloud and Abdelazim, Mona},
  editor = {El-Haj, Mo and Rayson, Paul  and Jarrar, Mustafa  and Ezeani, Ignatius  and Ezzini, Saad  and Ahmadi, Sina and Haddad Haddad, Amal and Amol, Cynthia and Abdelali, Ahmad and Abudalfa, Shadi},
  booktitle = {Proceedings of the 2nd Workshop on NLP for Languages Using Arabic Script},
  month     = mar,
  year      = {2026},
  address   = {Rabat, Morocco},
  publisher = {Association for Computational Linguistics},
  pages     = {313--321},
  doi       = {10.18653/v1/2026.abjadnlp-1.39},
  url       = {https://aclanthology.org/2026.abjadnlp-1.39/}
}

@article{kalajdzievski2023rslora,
  author  = {Damjan Kalajdzievski},
  title   = {A Rank Stabilization Scaling Factor for Fine-Tuning with {LoRA}},
  journal = {arXiv preprint arXiv:2312.03732},
  year    = {2023},
  url     = {https://arxiv.org/abs/2312.03732}
}

@inproceedings{liqreina-etal-2023-arabic,
  title     = {{A}rabic Fine-Grained Entity Recognition},
  author    = {Liqreina, Haneen Abdallatif and
               Jarrar, Mustafa and
               Khalilia, Mohammed and
               El-Shangiti, Ahmed Oumar and
               Abdul-Mageed, Muhammad},
  editor = {Sawaf, Hassan and El-Beltagy, Samhaa and Zaghouani, Wajdi and Magdy, Walid and Abdelali, Ahmed  and Tomeh, Nadi and Abu Farha, Ibrahim and Habash, Nizar and Khalifa, Salam  and Keleg, Amr and Haddad, Hatem and Zitouni, Imed and Mrini, Khalil and Almatham, Rawan},
  booktitle = {Proceedings of ArabicNLP 2023},
  month     = dec,
  year      = {2023},
  address   = {Singapore (Hybrid)},
  publisher = {Association for Computational Linguistics},
  pages     = {310--323},
  doi       = {10.18653/v1/2023.arabicnlp-1.25},
  url       = {https://aclanthology.org/2023.arabicnlp-1.25/}
}
\appendix

\section{Data Processing and Artifacts}
\label{app:relation-data}

\subsection{Source Resources and Processed Schema}

The shared-task data are derived from \wojoodrelations{}, which extends Wojood
with 40 directed relation types
\citep{aljabari-etal-2025-wojoodrelations}. Wojood supplies nested entity
annotations over 21 types \citep{jarrar-etal-2022-wojood}, while
WojoodOntology supplies relation definitions and directed domain--range
constraints \citep{aljabari-etal-2025-wojoodontology}.

We retained the original sentence, subject, object, and relation fields
unchanged and added occurrence-specific character offsets, token offsets,
and entity types. These additions support typed marker insertion, span validation, and ontology compatibility checks. 

\subsection{Verified Raw and Enriched Example}

Table~\ref{tab:raw-relation-example} shows a verified row from the original
\wojoodrelations{} training file. The row illustrates the task input before our
span and type enrichment.

\begin{table*}[t]
\centering
\small
\begin{tabularx}{\textwidth}{@{}lX@{}}
\toprule
\textbf{Field} & \textbf{Stored value} \\
\midrule
Sentence ID & \texttt{100002243} \\
Triple ID & \texttt{3830} \\
Sentence &
\textarabic{بلدية قلقيلية : الى قائمقام طولكرم / مدير منطقة قلقيلية :
قرارات مجلس محلي قلقيلية : بتاريخ ( 2 / 1 / 1964 ) .} \\
Subject & \textarabic{بلدية قلقيلية} \\
Object & \textarabic{قلقيلية} \\
Relation & \rel{Location.located\_in} \\
\bottomrule
\end{tabularx}
\caption{Verified example from the original \wojoodrelations{} training
artifact before our enrichment.}
\label{tab:raw-relation-example}
\end{table*}

The corresponding processed row retains all original fields and adds the
resolved occurrence information shown in
Table~\ref{tab:enriched-relation-example}. This example is particularly useful
because the object occurs inside the subject span, illustrating a nested
mention case.

\begin{table*}[t]
\centering
\small
\begin{tabularx}{\textwidth}{@{}lXX@{}}
\toprule
\textbf{Entity} & \textbf{Subject} & \textbf{Object} \\
\midrule
Surface form &
\textarabic{بلدية قلقيلية} &
\textarabic{قلقيلية} \\
Character span & $(0,13)$ & $(6,13)$ \\
Token span & $(0,1)$ & $(1,1)$ \\
Entity type & \texttt{ORG} & \texttt{GPE} \\
\bottomrule
\end{tabularx}

\vspace{2mm}
\caption{Occurrence-specific information added to the verified relation row
in Table~\ref{tab:raw-relation-example}. Offsets are reproduced as stored in
the processed artifact.}
\label{tab:enriched-relation-example}
\end{table*}

\subsection{Artifacts and Processing Stages}

\begin{table}[t]
\centering
\small
\begin{tabular}{lr}
\toprule
\textbf{Processing stage} & \textbf{Rows} \\
\midrule
Raw relation rows & 17,381 \\
After conflict removal & 17,208 \\
After ontology alignment and deduplication & 15,686 \\
Development artifact & 1,687 \\
Blind official input & 2,074 \\
\bottomrule
\end{tabular}
\caption{Relation artifact sizes and preprocessing stages.}
\label{tab:preprocessing-counts}
\end{table}

The blind input contained no gold relation labels. It was used only to produce
official predictions; we did not manually inspect, infer, or reconstruct its
hidden labels.

\subsection{Span and Entity-Type Recovery}

Training relation rows provide entity surface forms but not the
occurrence-specific spans and types required by our systems. We first align
mentions with Wojood flat and nested NER annotations
\citep{jarrar-etal-2022-wojood}, preferring exact span matches and allowing
compatible containment matches when necessary. Unresolved mentions are then
matched against predictions from the Wojood NER model. After these automatic
steps, two authors manually resolved the spans and entity types of the remaining
240 training rows, while retaining their original relation labels.

No manual span or type annotation is used for the blind set; its occurrences
and entity types are determined entirely by the inference pipeline described in
Appendix~\ref{app:span-resolution}.

\subsection{Span Validation}

For every processed row, we verified that the character slice identified by
the subject offsets reproduced the subject string and that the object slice
reproduced the object string. Rows with mismatched offsets were rejected rather
than silently accepted. The final processed training and development artifacts
reported zero slice errors.

This check is important because selecting the wrong repeated occurrence can
associate a relation label with unrelated textual evidence. At blind inference,
the systems resolve possible occurrence pairs as described in
Appendix~\ref{app:span-resolution}.

\subsection{Conflict Removal and Deduplication}

We first removed groups containing contradictory annotations for the same
resolved sentence and entity pair, reducing 17,381 raw rows to 17,208
non-conflicting rows. We then removed exact duplicates sharing the sentence,
subject, object, and relation label. Together
with the ontology-compatibility filtering described below, these preprocessing
steps produced 15,686 final training instances.
\subsection{Ontology Compatibility and Operational Mapping}

For each positive training instance, we checked the resolved subject and object
types against the directed domain--range constraints defined by WojoodOntology
\citep{aljabari-etal-2025-wojoodontology}. Because the task is directional,
compatibility is evaluated specifically for the ordered pair
$(\tau_s,\tau_o)$. Training rows whose subject--object type pair was not a
valid domain--range combination for the labeled relation were removed during
preprocessing. This prevents the models from being trained on relation
instances that are incompatible with the ontology under the resolved entity
types.

We made one systematic modification to the operational compatibility map based
on a highly frequent pattern observed in the labeled training data. In the
original WojoodOntology constraints, the domain and range of
\rel{Location.located\_in} are more restricted. However, the relation data
repeatedly contain labeled \rel{Location.located\_in} examples involving
\textsc{Org}, \textsc{Fac}, \textsc{Loc}, and \textsc{Gpe} entities in both
subject and object positions. We therefore expanded the operational
domain--range set for this relation to

\[
\begin{aligned}
r &= \rel{Location.located\_in},\\
\operatorname{Domain}(r)
  &= \{\textsc{Org}, \textsc{Fac}, \textsc{Loc}, \textsc{Gpe}\},\\
\operatorname{Range}(r)
  &= \{\textsc{Org}, \textsc{Fac}, \textsc{Loc}, \textsc{Gpe}\}.
\end{aligned}
\]
This modification was motivated by the recurring annotation pattern in the
provided relation data and was applied consistently during preprocessing and
ontology-based candidate filtering. It is an implementation-level adaptation
used by our system; it does not alter the published WojoodOntology itself or
the relation labels provided by the shared task.

\subsection{Relation and Entity-Type Distributions}

\begin{table}[t]
\centering
\small
\begin{tabular}{lr}
\toprule
\textbf{Relation} & \textbf{Training rows} \\
\midrule
\norel & 6,677 \\
\rel{Location.located\_in} & 4,269 \\
\rel{PartOf.geopolitical\_division} & 1,684 \\
\rel{Location.lives\_in} & 876 \\
\rel{Personal.has\_occupation} & 620 \\
\rel{Location.headquartered\_in} & 172 \\
\bottomrule
\end{tabular}
\caption{Selected frequent and difficult relations in the processed training
artifact.}
\label{tab:relation-distribution-app}
\end{table}

The subject side is dominated by organizations, persons, and geopolitical
entities, while the object side is dominated by geopolitical entities,
organizations, occupations, and locations. Frequent directed type pairs
therefore include organization--place, place--place,
organization--organization, person--place, and person--occupation.

\section{TypePredictor}
\label{app:typepredictor}

\subsection{Dataset Construction}

The TypePrediction collection was curated from the Wojood and WojoodFine NER
resources \citep{jarrar-etal-2022-wojood,liqreina-etal-2023-arabic}. Each example contains an Arabic sentence, one exact mention occurrence, and
its entity type. TypePredictor is trained only on entity-type labels and never
uses relation labels. Relation-extraction rows were excluded from this
curation. Exact duplicate mentions were collapsed, and same-span examples
with conflicting labels were removed.

The resulting collection contains 125,996 examples over 21 types:
100,796 for training, 12,600 for validation, and 12,600 for testing. The split
is a shuffled, stratified, mention-level 80/10/10 split. It was produced using
two stratified \texttt{train\_test\_split} operations with
\texttt{random\_state=42}. The same sentence may occur in multiple splits with
different entity spans. Because the split is mention-level, sentence-level
overlap with relation-extraction inputs, including the blind evaluation input,
is possible. Such overlap may affect the entity-type predictions subsequently
used for typed markers and ontology filtering, but it does not expose relation
labels to TypePredictor. Validation and test examples are therefore additionally
categorized as \texttt{seen\_sentence\_new\_entity} or
\texttt{unseen\_sentence}.
\begin{table*}[t]
\centering
\small
\begin{tabular}{rrrrrrrr}
\toprule
GPE & ORG & DATE & PERS & NORP & ORDINAL & OCC & EVENT \\
26,902 & 26,593 & 19,762 & 10,712 & 10,217 & 7,807 & 7,696 & 3,769 \\
\midrule
CARDINAL & LOC & WEBSITE & FAC & LAW & TIME & MONEY & CURR \\
3,724 & 2,388 & 1,478 & 1,296 & 904 & 871 & 420 & 411 \\
\midrule
LANGUAGE & PERCENT & PRODUCT & QUANTITY & UNIT & & & \\
332 & 312 & 190 & 106 & 106 & & & \\
\bottomrule
\end{tabular}
\caption{TypePrediction mention counts over all splits.}
\label{tab:type-counts}
\end{table*}

\subsection{Verified Example}

Table~\ref{tab:typeprediction-example} gives a verified TypePrediction training
row. Unlike the relation artifacts, the TypePrediction files do not physically
store the marked model input. The marked input is constructed at loading time from the sentence and stored character offsets.

\begin{table*}[t]
\centering
\small
\begin{tabularx}{\textwidth}{@{}lX@{}}
\toprule
\textbf{Field} & \textbf{Stored value} \\
\midrule
ID (\texttt{id}) & \texttt{tp\_093924} \\
Entity (\texttt{entity}) & \textarabic{المنظمة} \\
Type (\texttt{type}) & \texttt{ORG} \\
Sentence (\texttt{sentence}) &
\textarabic{كما تدير المنظمة مركزاً للإصابات البالغة في مديرية الحوبان ،
وهو يعالج جرحى الحرب والمرضى الذين يحتاجون إلى الرعاية الطبية . ,} \\
Token span (\texttt{start\_token}, \texttt{end\_token}) & $(2,2)$ \\
Character span (\texttt{start\_char}, \texttt{end\_char}) & $(9,16)$ \\
Raw types (\texttt{raw\_types}) & \texttt{ORG}, \texttt{NONGOV} \\
Sources (\texttt{sources}) &
\texttt{Wojood1\_1\_flat/test};
\texttt{Wojood1\_1\_nested/test};
\texttt{WojoodFine-Flat/split20} \\
\bottomrule
\end{tabularx}
\caption{Verified TypePrediction training example.}
\label{tab:typeprediction-example}
\end{table*}

Training rows store \texttt{entity}, \texttt{type}, \texttt{sentence},
\texttt{start\_token}, \texttt{end\_token}, \texttt{start\_char},
\texttt{end\_char}, \texttt{raw\_types}, \texttt{sources}, and \texttt{id};
validation and test rows additionally store \texttt{evaluation\_category}.
The marked input itself is not stored.

For the example in Table~\ref{tab:typeprediction-example}, the generated input
contains the fragment
\begin{quote}
\begin{Arabic}
كما تدير \LR{\texttt{[ENT]}} المنظمة \LR{\texttt{[/ENT]}} مركزاً للإصابات
البالغة في مديرية الحوبان، وهو يعالج جرحى الحرب والمرضى الذين يحتاجون إلى
الرعاية الطبية.
\end{Arabic}
\end{quote}

The model input is generated as
\[
x' =
x_{<s}\;
\texttt{[ENT]}\;
x_{s:e}\;
\texttt{[/ENT]}\;
x_{>e},
\]
possibly within a selected context window. The tokens \texttt{[ENT]} and
\texttt{[/ENT]} were explicitly added to the tokenizer as special tokens with
IDs 65,000 and 65,001, respectively, and the encoder embedding matrix was
resized to a vocabulary size of 65,002.

\subsection{Architecture and Training}

The TypePredictor consists of \texttt{U4RASD/NeoAraBERT} and a single
classification layer. Let $H\in\mathbb{R}^{L\times768}$ be the final hidden
states and let $\mathbf{h}_{\mathrm{cls}}=H_0$. The classifier computes
\begin{equation}
\begin{aligned}
\mathbf{z}^{\mathrm{type}}
&=
W_t\operatorname{Dropout}_{0.1}(\mathbf{h}_{\mathrm{cls}})
+\mathbf{b}_t,\\
\mathbf{z}^{\mathrm{type}}
&\in\mathbb{R}^{21}.
\end{aligned}
\end{equation}
The $768\rightarrow21$ head contains 16,149 parameters. Training minimizes
unweighted multiclass cross-entropy and inference returns the highest-scoring
type.

\begin{table}[t]
\centering
\small
\begin{tabular}{ll}
\toprule
\textbf{Setting} & \textbf{Value} \\
\midrule
Encoder & \texttt{U4RASD/NeoAraBERT} \\
Classes / maximum length & 21 / 512 \\
Epochs & 4 \\
Train/eval batch & 4 / 8 \\
Gradient accumulation & 4 \\
Encoder/head LR & $10^{-5}$ / $5\times10^{-5}$ \\
Optimizer & AdamW (\texttt{adamw\_torch}) \\
Scheduler & Linear \\
Weight decay / warm-up & 0.01 / 0.10 \\
Maximum gradient norm & 1.0 \\
Precision / seed & FP16 / 42 \\
Loss & Unweighted cross-entropy \\
Selection metric & Validation macro-$F_1$ \\
\bottomrule
\end{tabular}
\caption{TypePredictor training configuration.}
\label{tab:type-config}
\end{table}

\subsection{Detailed Evaluation}
\label{app:typepredictor-results}
\begin{figure*}[t]
\centering
\begin{minipage}{0.32\linewidth}
  \centering
  \includegraphics[width=\linewidth]{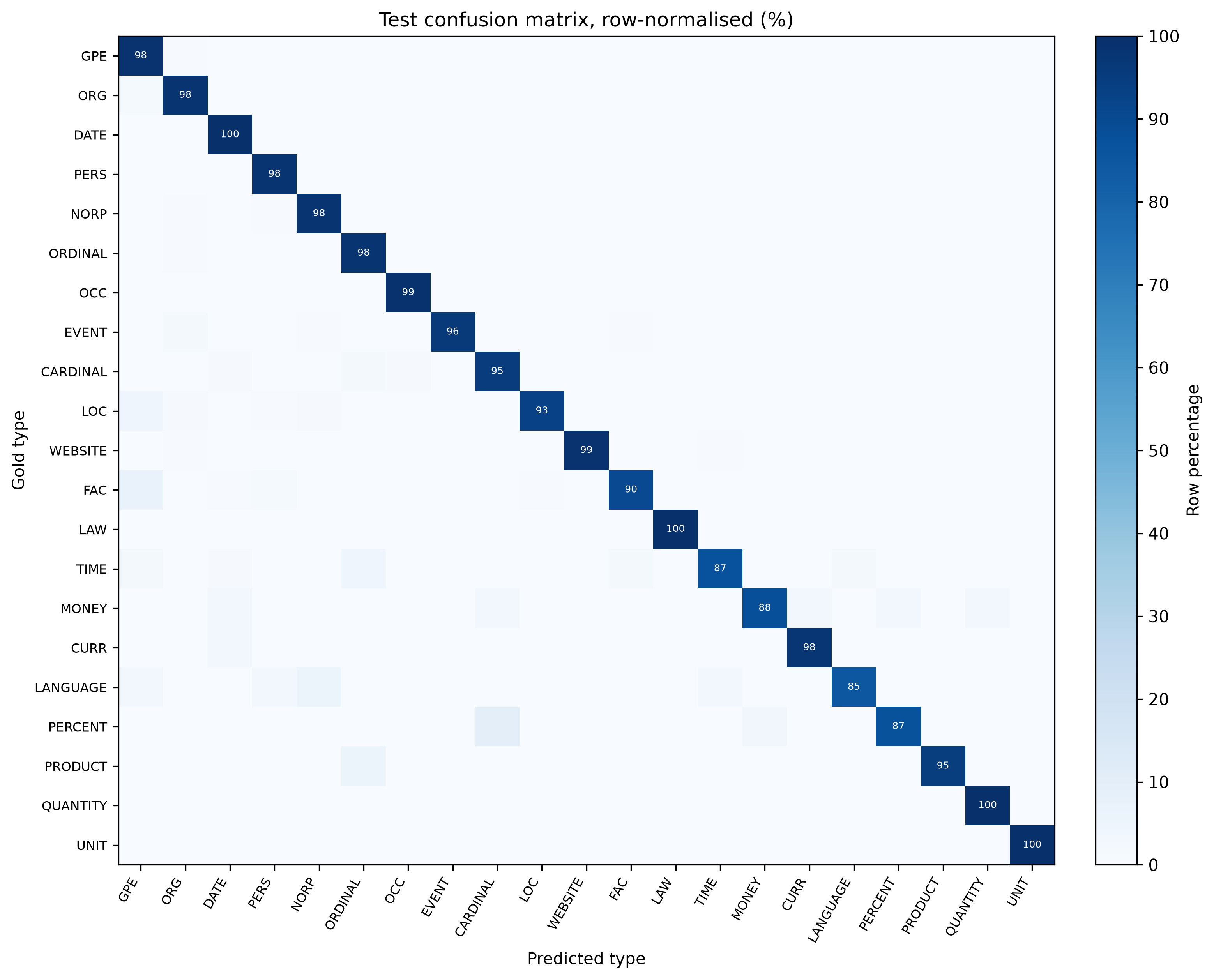}
  \small (a) Row-normalized confusion matrix.
\end{minipage}
\hfill
\begin{minipage}{0.32\linewidth}
  \centering
  \includegraphics[width=\linewidth]{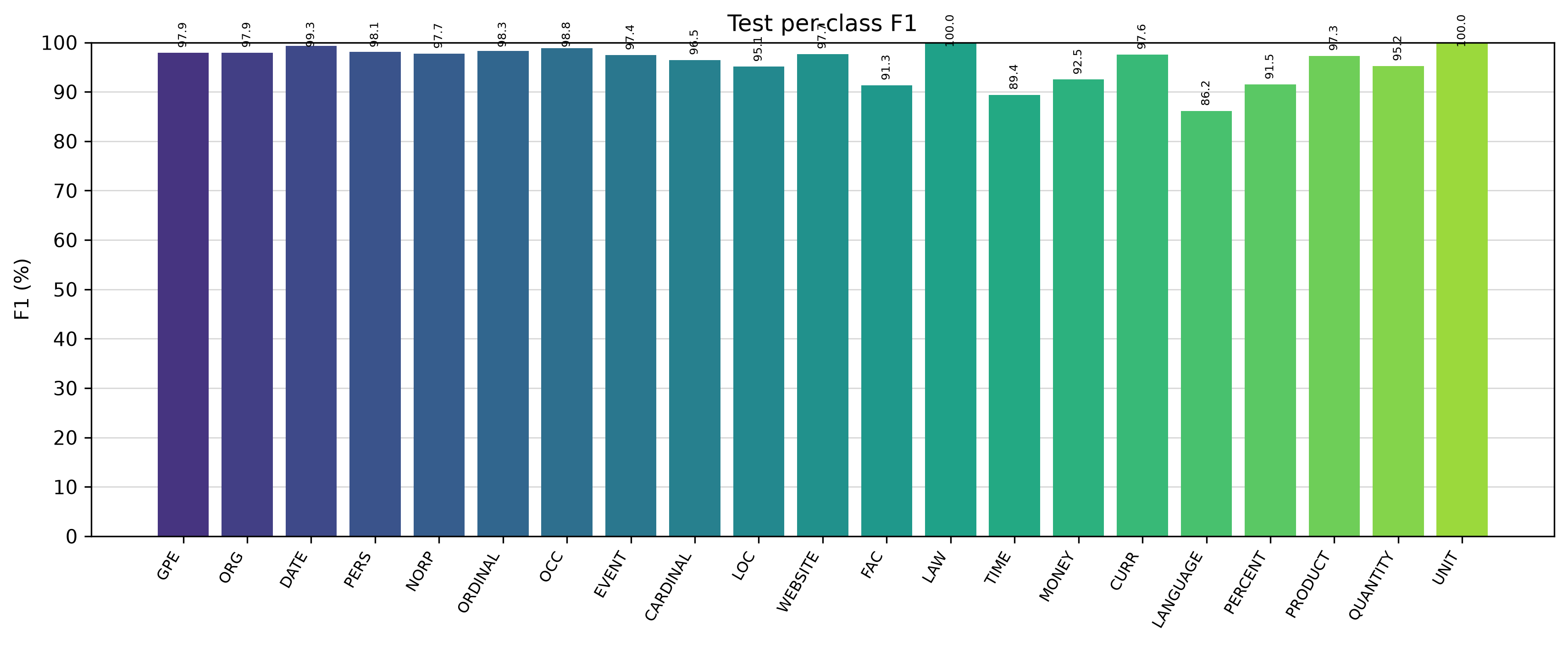}
  \small (b) Per-class test $F_1$.
\end{minipage}
\hfill
\begin{minipage}{0.32\linewidth}
  \centering
  \includegraphics[width=\linewidth]{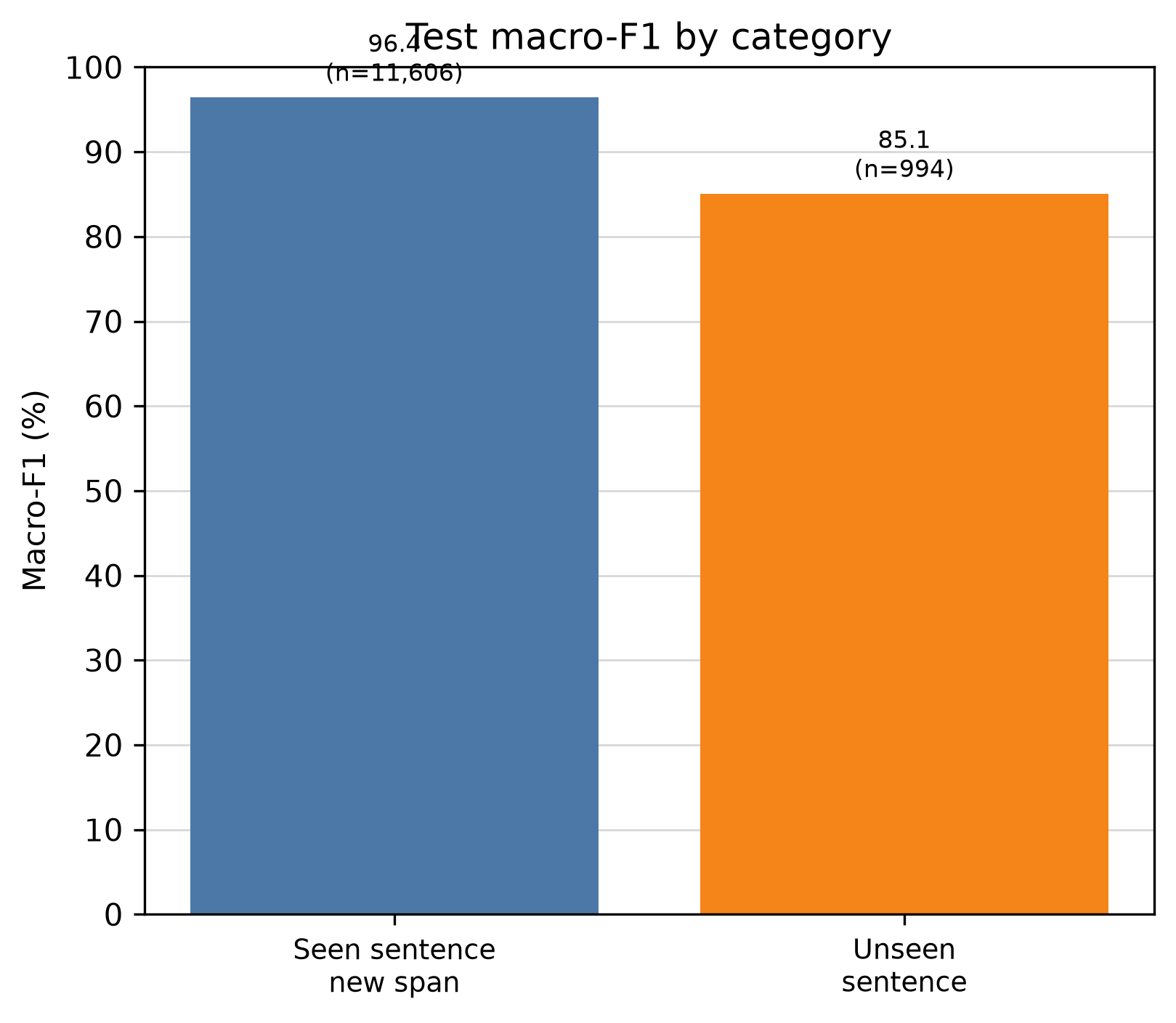}
  \small (c) Macro-$F_1$ by sentence category.
\end{minipage}
\caption{TypePredictor diagnostic plots on the held-out test split.}
\label{fig:typepredictor-diagnostics}
\end{figure*}

We further evaluate the TypePredictor using aggregate performance metrics,
sentence-level generalization categories, per-type performance, and diagnostic
visualizations. Overall validation and test performance is reported in
Table~\ref{tab:typepredictor-main-results}, while
Table~\ref{tab:typepredictor-category-results} separates performance between
seen sentences with new entity spans and entirely unseen sentences.
Table~\ref{tab:typepredictor-per-type-results} provides the detailed
per-type precision, recall, and $F_1$ scores on the test split.
Finally, Figure~\ref{fig:typepredictor-diagnostics} presents the normalized
confusion matrix, per-class $F_1$ scores, and performance across sentence
categories.

\begin{table*}[t]
\centering
\small
\begin{tabular}{lrrrrrrr}
\toprule
\textbf{Split} & \textbf{Rows} & \textbf{Accuracy} &
\textbf{Micro-$F_1$} & \textbf{Macro-$F_1$} &
\textbf{Weighted-$F_1$} & \textbf{Top-3} & \textbf{Top-5} \\
\midrule
Validation & 12,600 & 97.79 & 97.79 & 95.87 & 97.78 & 99.36 & 99.64 \\
Test & 12,600 & 97.94 & 97.94 & 95.99 & 97.93 & 99.35 & 99.62 \\
\bottomrule
\end{tabular}
\caption{TypePredictor performance. Scores are percentages.}
\label{tab:typepredictor-main-results}
\end{table*}

\begin{table*}[t]
\centering
\small
\begin{tabular}{llrrrrr}
\toprule
\textbf{Split} & \textbf{Category} & \textbf{Rows} &
\textbf{Accuracy} & \textbf{Macro-$F_1$} &
\textbf{Weighted-$F_1$} & \textbf{Top-5} \\
\midrule
Validation & Seen sentence/new span & 11,661 & 98.04 & 96.33 & 98.04 & 99.70 \\
Validation & Unseen sentence & 939 & 94.57 & 86.56 & 94.50 & 98.94 \\
Test & Seen sentence/new span & 11,606 & 98.08 & 96.41 & 98.07 & 99.67 \\
Test & Unseen sentence & 994 & 96.28 & 85.06 & 96.18 & 98.99 \\
\bottomrule
\end{tabular}
\caption{TypePredictor results by sentence category.}
\label{tab:typepredictor-category-results}
\end{table*}

\begin{table}[t]
\centering
\scriptsize
\begin{tabular}{lrrrr}
\toprule
\textbf{Type} & \textbf{Support} & \textbf{P} & \textbf{R} & \textbf{$F_1$} \\
\midrule
GPE & 2,691 & 97.39 & 98.44 & 97.91 \\
ORG & 2,659 & 98.08 & 97.82 & 97.95 \\
DATE & 1,977 & 99.04 & 99.54 & 99.29 \\
PERS & 1,071 & 97.95 & 98.32 & 98.14 \\
NORP & 1,021 & 97.65 & 97.85 & 97.75 \\
ORDINAL & 781 & 98.21 & 98.34 & 98.27 \\
OCC & 770 & 98.70 & 98.96 & 98.83 \\
EVENT & 377 & 98.64 & 96.29 & 97.45 \\
CARDINAL & 372 & 97.79 & 95.16 & 96.46 \\
LOC & 239 & 96.96 & 93.31 & 95.10 \\
WEBSITE & 148 & 96.69 & 98.65 & 97.66 \\
FAC & 129 & 92.80 & 89.92 & 91.34 \\
LAW & 91 & 100.00 & 100.00 & 100.00 \\
TIME & 87 & 91.57 & 87.36 & 89.41 \\
MONEY & 42 & 97.37 & 88.10 & 92.50 \\
CURR & 41 & 97.56 & 97.56 & 97.56 \\
LANGUAGE & 33 & 87.50 & 84.85 & 86.15 \\
PERCENT & 31 & 96.43 & 87.10 & 91.53 \\
PRODUCT & 19 & 100.00 & 94.74 & 97.30 \\
QUANTITY & 10 & 90.91 & 100.00 & 95.24 \\
UNIT & 11 & 100.00 & 100.00 & 100.00 \\
\bottomrule
\end{tabular}
\caption{Per-type TypePredictor results on the 12,600-row test split.}
\label{tab:typepredictor-per-type-results}
\end{table}

\section{Shared Inference Pipeline}
\label{app:span-resolution}

\subsection{Marker Insertion and Overlap Handling}

Markers are inserted from character spans rather than through global surface
replacement. Insertions are ordered to keep nested, overlapping, adjacent, and
equal-span cases deterministic. After tokenization, the system verifies that
the required opening and closing marker IDs remain present. When necessary,
the sentence is cropped around the outer entity boundaries.

\subsection{Ontology Filtering}

For a predicted directed type pair $(\tau_s,\tau_o)$, the operational
WojoodOntology map returns the compatible positive relations. \norel{} remains
available for every pair. Incompatible encoder and classifier logits are
masked, while generative prompts display or score only compatible labels.

\subsection{Candidate-Pair Handling}

The model families use different strategies when several occurrence pairs are
possible; there is no single nearest-distance rule shared by all released
runners. TypedPair can score all candidate pairs and aggregate their
predictions as described in Appendix~\ref{app:TypedPair}. Yehia-RE also scores
candidate pairs and applies the row-level aggregation rules described in
Appendix~\ref{app:yehia}.

For Qwen-RE configurations using the nearest-candidate heuristic, let the
subject span be $[s_s,s_e)$ and the object span be $[o_s,o_e)$. Candidate pairs
are sorted lexicographically by

\begin{enumerate}[leftmargin=*]
\item character gap
\[
d_{\mathrm{gap}}
=
\max\!\left(0,
\max(s_s,o_s)-\min(s_e,o_e)
\right);
\]
\item twice the absolute distance between span centers,
\[
d_{\mathrm{center}}
=
|s_s+s_e-o_s-o_e|;
\]
\item ascending subject start offset $s_s$; and
\item ascending object start offset $o_s$.
\end{enumerate}

The first candidate under this ordering is selected. Overlapping and nested
mentions therefore receive a character gap of zero, and the final two offsets
make ties deterministic.

\section{TypedPair Encoder Systems}
\label{app:TypedPair}

\subsection{Typed Directional Input}

For subject and object types $\tau_s$ and $\tau_o$, the sentence receives typed
opening and closing markers such as \texttt{[SUBJ\_PERS]} and
\texttt{[OBJ\_ORG]}. Across 21 entity types, this introduces 84 marker strings.
We additionally construct the directional type vector
\begin{equation}
\mathbf{t}_{s,o}
=
[
\operatorname{onehot}(\tau_s);
\operatorname{onehot}(\tau_o)
]
\in\mathbb{R}^{42},
\end{equation}
where the first 21 coordinates always represent the subject and the final
21 represent the object.

\subsection{Pair Representation and Two-Head Objective}

Let $H\in\mathbb{R}^{L\times d}$ be the encoder output, and let
$\mathbf{h}_c$, $\mathbf{h}_s$, and $\mathbf{h}_o$ denote the sequence-start,
subject-opening, and object-opening states. The directed pair representation is
\begin{equation}
\mathbf{u}_{s,o}
=
[
\mathbf{h}_c;
\mathbf{h}_s;
\mathbf{h}_o;
|\mathbf{h}_s-\mathbf{h}_o|;
\mathbf{h}_s\odot\mathbf{h}_o;
\mathbf{t}_{s,o}
].
\end{equation}
Its dimensionality is $5d+42$. A shared MLP applies
$(5d+42)\rightarrow1024\rightarrow512$ using LayerNorm, GELU, and dropout.

The existence head returns
$\mathbf{z}^{e}\in\mathbb{R}^{2}$, while the positive-relation head returns
$\mathbf{z}^{+}\in\mathbb{R}^{40}$. For relation label $y$, define
$a=\mathbf{1}[y\neq\norel]$. Training minimizes
\begin{equation}
\mathcal{L}_{\mathrm{EP}}
=
\operatorname{CE}(\mathbf{z}^{e},a)
+
a\,\operatorname{CE}(\mathbf{z}^{+},y^+).
\end{equation}
Thus, every row supervises relation existence, whereas negative rows do not
assign an arbitrary target to the positive-relation head.

\subsection{Row-Level Candidate Strategies}

For occurrence-pair candidate $c$, let
\begin{equation}
\begin{aligned}
p_c &= P_c(\text{relation exists}),\\
q_c(r) &= P_c(r \mid \text{relation exists}).
\end{aligned}
\end{equation}
after ontology-incompatible positive relations have been masked. The released
pipeline contains several deterministic alternatives for combining predictions
across occurrence pairs.

\paragraph{Majority + logit-sum.}
Let $P=\{c:p_c\geq\tau\}$ among $N$ candidates. The row is treated as positive
only when
\[
|P|\geq\left\lceil0.5N\right\rceil.
\]
For a positive row, masked positive-relation logits are summed over the passing
candidates and the relation with the largest aggregate score is selected.

\paragraph{Top existence.}
First select
\[
c^*=\arg\max_c p_c.
\]
If $p_{c^*}<\tau$, the row is assigned \norel{}. Otherwise, the output is
\[
\arg\max_r q_{c^*}(r).
\]

\paragraph{Max joint.}
Among candidates that pass the existence threshold, select
\begin{equation}
(c^*,r^*)
=
\arg\max_{c,r}
p_c q_c(r).
\end{equation}
If no candidate passes the threshold, the result is \norel{}.

\paragraph{Soft joint pooling.}
Positive evidence for relation $r$ is pooled as
\begin{equation}
S(r)
=
\sum_{c:p_c\geq\tau}
p_c q_c(r),
\end{equation}
and the relation with the largest pooled score is returned when a positive
decision is made. Ties are resolved deterministically using decision score,
existence probability, candidate order, and relation index.

\subsection{Dimensions and Training}

\begin{table}[t]
\centering
\small
\begin{tabular}{lll}
\toprule
\textbf{Operation} & \textbf{Base} & \textbf{Large} \\
\midrule
Encoder states & $L\times768$ & $L\times1024$ \\
Pair vector & 3,882 & 5,162 \\
Shared hidden & 512 & 512 \\
Existence logits & 2 & 2 \\
Positive logits & 40 & 40 \\
\bottomrule
\end{tabular}
\caption{TypedPair tensor dimensions.}
\label{tab:TypedPair-shapes}
\end{table}

For $d=768$, the MLP and heads add 4,525,610 parameters; for $d=1024$, they
add 5,836,330. The no-interactions variant uses
$[\mathbf{h}_c;\mathbf{h}_s;\mathbf{h}_o;\mathbf{t}_{s,o}]$ and adds
2,952,746 parameters.

Training uses four epochs, maximum length 512, physical batch size 4,
gradient accumulation 4, encoder learning rate $10^{-5}$, new-layer learning
rate $5\times10^{-5}$, weight decay 0.01, 10\% warm-up, FP16, and MLP dropout
0.1. Evaluation occurs every 250 optimizer steps. For the released
AraModernBERT model, the existence threshold selected on labeled validation
data is 0.01, with positive micro-$F_1$ used for threshold selection.

\subsection{Extended Official Results}

Table~\ref{tab:TypedPair-results-app} summarizes the official scores across
encoder configurations, with AraModernBERT obtaining the highest reported
$F_1$ score of 92.69.

\label{app:TypedPair-results}

\begin{table}[t]
\centering
\small
\begin{tabularx}{\columnwidth}{@{}Xr@{}}
\toprule
\textbf{Configuration} & \textbf{Official $F_1$} \\
\midrule
NeoAraBERT, interactions, majority + logit-sum & 88.33 \\
NeoAraBERT, top existence & 88.59 \\
AraBERTv02 & 89.33 \\
NeoAraBERT, no interactions & 89.36 \\
ARBERTv2 & 91.96 \\
AraBERTv02-Large, soft pool & 92.20 \\
AraBERTv02-Large, top existence & 92.29 \\
AraModernBERT & \textbf{92.69} \\
\bottomrule
\end{tabularx}
\caption{Official TypedPair results.}
\label{tab:TypedPair-results-app}
\end{table}

\subsection{AraModernBERT Component Ablation}
\label{app:TypedPair-ablation}

AraModernBERT achieved the strongest encoder result and was therefore used to
compare configurations with and without typed text markers and the directional
type vector.
Table~\ref{tab:aramodern-ablation} reports the corresponding official
full-pipeline ablation submissions.

\begin{table}[t]
\centering
\small
\begin{tabular}{ccc}
\toprule
\textbf{Typed markers} & \textbf{Type vector} & \textbf{Official $F_1$} \\
\midrule
No & No & 90.65 \\
No & Yes & 90.75 \\
Yes & No & 89.94 \\
Yes & Yes & \textbf{92.69} \\
\bottomrule
\end{tabular}
\caption{AraModernBERT component ablation on the blind evaluation.}
\label{tab:aramodern-ablation}
\end{table}

The complete configuration obtains the strongest result among these four
submissions. The ablation artifacts share the intended four-epoch training
setup, batch sizes, learning rates, weight decay, warm-up, MLP dimensions,
dropout, and TypePredictor dependency.

\section{Yehia-RE}
\label{app:yehia}
\subsection{Row-Specific Arabic Multiple Choice}

Early Yehia variants generated full Arabic relation phrases and were vulnerable
to explanatory text, punctuation, and harmless lexical variation. Yehia-RE
instead reformulates each instance as a row-specific Arabic multiple-choice
problem.

For each ontology-compatible relation, the option shown to the model is an
Arabic natural-language template (\texttt{template\_ar}), rather than the
canonical relation identifier. Our Arabic relation templates were inspired by
the relation-aware templates used in the \wojoodrelations{}/SinaTools
implementation \citep{aljabari-etal-2025-wojoodrelations}.\footnote{
\url{https://github.com/SinaLab/sinatools}}
Thus, labels such as \rel{Affiliation.employee\_of} are used internally to
construct and evaluate the example, while the model sees the corresponding
Arabic formulation, such as \textarabic{الأول يعمل لدى الثاني}. The canonical
relation names are not presented as answer options.

Each displayed Arabic option is assigned a verified one-token Arabic code.
The available codes are
\textarabic{أ، ب، ج، د، ه، و، ز، ح، ط، ي}.
The codes are row-local addresses rather than global relation-class IDs, and
\norel{} is represented by its own Arabic option
\textarabic{لا توجد علاقة}. The largest ontology-compatible candidate set contains nine positive relations,
for the directed type pair \textsc{Pers}$\rightarrow$\textsc{Org}. Including
\norel{}, a row therefore contains at most ten options, matching the ten
available Arabic answer codes.

\subsection{Full-Vocabulary and Row-Local Objectives}

In both variants, all prompt and padding tokens are masked from the training
loss so that exactly one assistant code token is supervised.

For the full-vocabulary variant, let $g$ be the gold code and
$\mathcal{V}$ the complete model vocabulary. The objective is
\begin{equation}
\mathcal{L}_{\mathrm{FV}}
=
-\log
\frac{\exp z_g}
{\sum_{v\in\mathcal{V}}\exp z_v}.
\end{equation}
At inference, only the codes corresponding to options displayed in the current
row are compared.

The row-local variant instead normalizes directly over the displayed code set
$C_i$:
\begin{equation}
\mathcal{L}_{\mathrm{RL}}
=
-\log
\frac{\exp z_g}
{\sum_{c\in C_i}\exp z_c}.
\end{equation}

\subsection{Prompt Template}

The Yehia-RE system message is:

\begin{quote}
\begin{Arabic}
أنت مصنّف علاقات. حدّد العلاقة الموجّهة من الكيان [الأول] إلى الكيان
[الثاني].\\
اختر علاقة واحدة من القائمة تناسب معنى الجملة ونوعي الكيانين.\\
إن لم تنطبق أي علاقة فاختر «لا توجد علاقة».\\
أجب برمز الخيار فقط: حرف عربي واحد دون رقم أو شرح.
\end{Arabic}
\end{quote}

The row-specific user message follows this structure:

\begin{quote}
\begin{Arabic}
الجملة:\\
\{marked\_sentence\}\\[3pt]

نوع الأول: \{first\_type\_ar\}\\
نوع الثاني: \{second\_type\_ar\}\\[3pt]

الخيارات:\\
أ) \{option 1\}\\
ب) \{option 2\}\\
\ldots\\
\{last code\}) لا توجد علاقة\\[3pt]

رمز الإجابة:
\end{Arabic}
\end{quote}

The assistant message contains exactly one displayed code.
\subsection{Complete Training Example}

The following is a fully instantiated training example from the released
record \texttt{train\_1831}. The canonical relation identifiers are used only
internally to construct and evaluate the example; the actual prompt presents
the corresponding Arabic \texttt{template\_ar} options.

\paragraph{System message.}
\begin{quote}
\begin{Arabic}
أنت مصنّف علاقات. حدّد العلاقة الموجّهة من الكيان [الأول] إلى الكيان
[الثاني].\\
اختر علاقة واحدة من القائمة تناسب معنى الجملة ونوعي الكيانين.\\
إن لم تنطبق أي علاقة فاختر «لا توجد علاقة».\\
أجب برمز الخيار فقط: حرف عربي واحد دون رقم أو شرح.
\end{Arabic}
\end{quote}

\paragraph{User message.}
\begin{quote}
\begin{Arabic}
الجملة:\\
``نجدّد نداءاتنا للسلطات التركية المعنية لتسهيل العبور العاجل للإمدادات
الأساسية والموظفين إلى شمال غرب سوريا للسماح بتوسيع نطاق استجابتنا
الإنسانية والطبية في المنطقة.'' [الأول] كريستيان ريندرز [/الأول]،
منسّق أنشطة [الثاني] أطباء بلا حدود [/الثاني] في شمال غرب سوريا.\\[3pt]

نوع الأول: شخص\\
نوع الثاني: منظمة\\[3pt]

الخيارات:\\
أ) الأول قائد للثاني\\
ب) الأول مؤسس للثاني\\
ج) الأول رئيس للثاني\\
د) الأول مدير للثاني\\
ه) الأول عضو في الثاني\\
و) الأول طالب لدى الثاني\\
ز) الأول يعمل لدى الثاني\\
ح) الأول منافس للثاني\\
ط) الأول مالك للثاني\\
ي) لا توجد علاقة\\[3pt]

رمز الإجابة:
\end{Arabic}
\end{quote}

\paragraph{Assistant target.}
\begin{quote}
\begin{Arabic}
ز
\end{Arabic}
\end{quote}

The target code \textarabic{ز} maps internally to
\rel{Affiliation.employee\_of}. The canonical relation identifier is not shown
to the model as an answer option; instead, the model sees its Arabic
\texttt{template\_ar}, \textarabic{الأول يعمل لدى الثاني}. Only the single
assistant code token contributes to the training loss.
\subsection{Deterministic Transformation}

For each training row, the builder validates spans and types, inserts
\textarabic{[الأول]} and \textarabic{[الثاني]} with overlap-safe ordering,
maps types to Arabic names, retrieves compatible relations, renders their
Arabic \texttt{template\_ar} formulations, deterministically shuffles positive
options with seed 42,
appends \norel{}, assigns one-token codes, maps the gold relation to its code,
and builds the chat messages.

The corresponding audit checks marker occurrence, option uniqueness,
domain/range compatibility, code-token alignment, reversibility, maximum
sequence length, and absence of canonical relation-identifier leakage from the
answer.

\subsection{QLoRA Configuration}

\begin{table*}[t]
\centering
\small
\begin{tabular}{llll}
\toprule
\textbf{Category} & \textbf{Setting} & \textbf{Value} & \textbf{Notes} \\
\midrule
Base & Model & \texttt{Navid-AI/Yehia-7B-preview} & pinned revision \\
Base & Revision &
\multicolumn{2}{l}{\texttt{b9dda4715eafee7e8090d2c83cfe078d75f4ebb8}} \\
Runtime & Quantization & 4-bit NF4 + double quant. & BF16 compute \\
Runtime & Maximum length & 1,024 & gradient checkpointing \\
LoRA & Target & eligible internal linear layers & RS-LoRA \\
LoRA & $r/\alpha$/dropout & 16 / 32 / 0.05 & 39,976,960 trainable \\
Training & Batch/accumulation & 4 / 4 & effective batch 16 \\
Training & Epochs/steps & 3 / 2,943 & best step 2,750 \\
Training & LR / warm-up & $5\times10^{-5}$ / 5\% & cosine, zero decay \\
Training & Optimizer & \texttt{paged\_adamw\_8bit} & clipping 1.0 \\
Sampling & Row weight & $n_c^{-0.25}$ & replacement \\
Evaluation & Frequency & 250 steps & full development set \\
Seed & Data and model & 42 & one reported seed \\
\bottomrule
\end{tabular}
\caption{Selected Yehia-RE QLoRA configuration.}
\label{tab:yehia-qlora}
\end{table*}

Yehia-RE uses rank-stabilized LoRA (RS-LoRA)
\citep{kalajdzievski2023rslora}. The all-linear audit finds 225 supported
linear/head modules: seven internal projections in each of 32 transformer
layers plus the language-model head. PEFT wraps the 224 internal modules and
leaves \texttt{lm\_head} frozen. Only the 448 LoRA matrices receive gradients.

\subsection{One-Token Masking Audit}

For each labeled row, the system message, user message, chat-template boundary,
trailing template tokens, and padding positions receive label $-100$. Exactly
one assistant code token contributes to the loss. A row is rejected if the
prompt/full-conversation boundary differs, the answer code is not one token,
or the sequence exceeds 1,024 tokens.

\subsection{Candidate-Pair Aggregation}

For occurrence-pair candidate $c$, let $\pi_c(k)$ denote the normalized
probability of displayed option code $k$. Let $k_c^*$ be the winning code,
$r_c$ its relation, and define
\[
u_c=\pi_c(k_c^*)
\]
as the candidate confidence. We also define the score margin
\[
m_c=
\ell_c^{(1)}-\ell_c^{(2)},
\]
where $\ell_c^{(1)}$ and $\ell_c^{(2)}$ are the largest and second-largest option log-probabilities.

The released runner implements the following row-level strategies.

\paragraph{Top confidence.}
Select from all occurrence-pair candidates according to the lexicographic key
\[
(u_c,m_c,-c),
\]
where $c$ is the deterministic candidate index. The winning candidate's
relation, including \norel{} when applicable, is returned.

\paragraph{Max positive.}
Discard candidates whose predicted relation is \norel{}. If no positive
candidate remains, return \norel{}. Otherwise, select the remaining candidate
using the same lexicographic ordering
\[
(u_c,m_c,-c)
\]
and return its predicted positive relation.

\paragraph{Majority vote.}
Let $N$ be the total number of candidate occurrence pairs and let
$\rho=0.5$. Positive voting is allowed only when at least
\[
\left\lceil\rho N\right\rceil
\]
candidates predict a positive relation; otherwise the row is assigned
\norel{}. Positive candidates vote by predicted relation. A relation-count tie
is broken by, in order:

\begin{enumerate}[leftmargin=*]
\item the sum of normalized probability support for the tied relation over
all candidates;
\item the largest winning-candidate margin for that relation; and
\item canonical relation order.
\end{enumerate}

\subsection{vLLM Inference}

The released blind-test runner uses BF16 bitsandbytes loading, one GPU,
\texttt{tensor\_parallel\_size=1}, maximum length 1,024, prefix caching,
temperature 0, top-$p=1$, one generated token, and an allowed-token list equal
to the codes displayed in the current row. Requests are batched up to 512
prompts.

\subsection{Representative Official Results}

Table~\ref{tab:yehia-results-app} summarizes the official results of the two
Yehia-RE training variants.

\begin{table}[t]
\centering
\small
\begin{tabular}{llr}
\toprule
\textbf{Variant} & \textbf{Row decision} & \textbf{Official $F_1$} \\
\midrule
Full-vocabulary & Max positive & \textbf{92.98} \\
Row-local & Max positive & 92.84 \\
\bottomrule
\end{tabular}
\caption{Representative official Yehia-RE results.}
\label{tab:yehia-results-app}
\end{table}

\section{Qwen-RE}
\label{app:qwen}

\subsection{Causal-LM Prompt Template}
\label{app:qwen-causal-prompt}

The Qwen-RE causal-LM prompt is constructed separately for each directed
subject--object pair. It contains a system message defining the relation
extraction task and required output format, followed by a user message
containing the marked Arabic sentence, the subject and object strings and
their predicted entity types, and only the relations compatible with the
directed type pair under the operational ontology. Each compatible relation is
presented with its label, Arabic template, direction note, and concise meaning. The generic prompt structure is shown below, where
braced expressions denote row-specific values.

\begin{quote}
\small
\textbf{System message}

You are an Arabic relation extraction system.

Given one Arabic sentence with a marked subject and object, return exactly one
relation label.

You must always choose one label from the allowed label set.

Use the Arabic relation templates and direction notes as ontology definitions.

Prefer a positive relation label whenever the sentence provides plausible
evidence for one.

Use \norel{} only as a last resort when none of the allowed positive relations
is supported.

Do not copy explanations, templates, or direction notes.

Return only the exact label text and nothing else.

\medskip
\textbf{User message}

Classify the relation from the marked SUBJECT to the marked OBJECT.

The sentence uses \texttt{[SUBJ/...][/SUBJ/...]} and
\texttt{[OBJ/...][/OBJ/...]} markers.

Use the Arabic ontology template beside each allowed label to understand its
meaning.

Follow each direction note strictly because relation direction changes the
label.

Choose exactly one label from the provided label set.

Prefer choosing the best-supported positive relation.

Use \norel{} only when none of the allowed positive relations fits the marked
sentence.

\medskip
\textbf{Task rules}

1. Read the Arabic sentence and identify the relation between the marked
subject and object.

2. Choose exactly one relation label.

3. Prefer the best-supported positive relation instead of defaulting to
\norel{}.

4. Use \norel{} only when none of the allowed positive relations fits the
marked sentence.

5. Use the Arabic templates and direction notes as label definitions, not as
text to copy.

6. Return only the label text with no explanation.

\medskip
\textbf{Allowed labels and ontology meanings for this subject--object type
pair}

Direction: classify the relation from SUBJECT to OBJECT.

\medskip
\rel{\{RelationLabel\}}: Arabic template:
\textarabic{\LR{SUBJECT} \{relation template\} \LR{OBJECT}}

Direction: \{relation-specific direction note\}

Meaning: \{concise relation meaning\}

\medskip
\textit{The preceding relation block is repeated for every
ontology-compatible positive relation.}

\medskip
\norel{}: Arabic template:
\textarabic{لا توجد علاقة صالحة بين \LR{SUBJECT} و \LR{OBJECT} في هذه الجملة}

Direction: use only when no allowed positive relation is supported by the
marked sentence.

Meaning: no allowed positive relation is supported between the marked SUBJECT
and OBJECT in the required direction.

\medskip
\textbf{Subject:} \textarabic{\{subject\}}

\textbf{Object:} \textarabic{\{object\}}

\textbf{Subject type:} \texttt{\{subject type\}}

\textbf{Object type:} \texttt{\{object type\}}

\medskip
\textbf{Sentence:}

\textarabic{\{sentence containing the typed SUBJECT and OBJECT markers\}}

\medskip
\textbf{Answer:}
\end{quote}

During supervised fine-tuning, the assistant response following this prompt
contains only the gold relation label. At inference, no gold label is
provided: the constructed prompt ends at \texttt{Answer:}, after which the
causal language model predicts the relation label. Because the candidate
inventory is derived from the predicted directed entity-type pair, the allowed
label set can differ from one input row to another.

\subsection{Verified Causal-LM Training Example}
\label{app:qwen-causal-example}

The following example instantiates the prompt in
Appendix~\ref{app:qwen-causal-prompt} using the verified training row
\texttt{triple\_id=3830} from Table~\ref{tab:raw-relation-example}.

\begin{quote}
\small

\textbf{Subject:} \textarabic{بلدية قلقيلية}

\textbf{Object:} \textarabic{قلقيلية}

\textbf{Subject type:} \texttt{ORG}

\textbf{Object type:} \texttt{GPE}

\medskip
\textbf{Allowed relations:}

\rel{Business.has\_conflict\_with}:
\textarabic{\LR{SUBJECT} لديه نزاع مع \LR{OBJECT}}

\rel{Location.headquartered\_in}:
\textarabic{\LR{SUBJECT} يقع مقره الرئيسي في \LR{OBJECT}}

\rel{Location.located\_in}:
\textarabic{\LR{SUBJECT} يقع في \LR{OBJECT}}

\rel{Location.nearby}:
\textarabic{\LR{SUBJECT} يقع بالقرب من \LR{OBJECT}}

\norel{}:
\textarabic{لا توجد علاقة صالحة بين \LR{SUBJECT} و \LR{OBJECT} في هذه الجملة}

\medskip
\textbf{Sentence:}

\textarabic{
\LR{\texttt{[SUBJ/ORG]}}
بلدية
\LR{\texttt{[OBJ/GPE]}}
قلقيلية
\LR{\texttt{[/OBJ/GPE]}}
\LR{\texttt{[/SUBJ/ORG]}}
: الى قائمقام طولكرم / مدير منطقة قلقيلية : قرارات مجلس محلي قلقيلية :
بتاريخ ( 2 / 1 / 1964 ) .
}

\medskip
\textbf{Assistant target:}
\rel{Location.located\_in}

\end{quote}

During inference, the same row-specific prompt content is used without the
assistant target, and the model predicts one of the allowed labels.

\subsection{Generation and Complete-Label Scoring}

Free-form generation can produce explanations, punctuation, partial labels, or
an incompatible continuation. We therefore evaluate complete-label
scoring. Let $\mathcal{C}(x)$ denote the ontology-compatible candidate set,
including \norel{}. For candidate label
$c=(y_1,\ldots,y_{T_c})$, its score is the mean token log-probability
\begin{equation}
s(c\mid x)
=
\frac{1}{T_c}
\sum_{i=1}^{T_c}
\log p_{\theta}(y_i\mid x,y_{<i}).
\end{equation}
The selected label is
\begin{equation}
\widehat c
=
\arg\max_{c\in\mathcal{C}(x)}
s(c\mid x).
\end{equation}

For analysis, scores can be normalized over the compatible set:
\begin{equation}
P(c\mid x,\mathcal{C})
=
\frac{\exp s(c\mid x)}
{\sum_{c'\in\mathcal{C}(x)}\exp s(c'\mid x)}.
\end{equation}
This normalization does not change the highest-scoring label, although its
value depends on the size and composition of the candidate set.

\subsection{Sequence-Classifier Prompt and Objective}
\label{app:qwen-classifier-prompt}

Qwen-RE-9B-Cls receives the same marked sentence and entity metadata but uses
a shorter instruction because the model learns a fixed 41-way classification
head.

A rendering for the same verified example is:

\begin{quote}
\small
You are an expert Arabic relation extraction encoder.

Determine the semantic relation from the marked SUBJECT to the marked OBJECT.
Respect relation direction and use \norel{} only when no positive relation is
stated or clearly implied. The classification head selects the final class; do
not generate an answer.

\medskip
\textbf{Arabic sentence:}

\textarabic{
\LR{\texttt{[SUBJ/ORG]}}
بلدية
\LR{\texttt{[OBJ/GPE]}}
قلقيلية
\LR{\texttt{[/OBJ/GPE]}}
\LR{\texttt{[/SUBJ/ORG]}}
: الى قائمقام طولكرم / مدير منطقة قلقيلية : قرارات مجلس محلي قلقيلية :
بتاريخ ( 2 / 1 / 1964 ) .
}

\medskip
\textbf{SUBJECT:} \textarabic{بلدية قلقيلية}

\textbf{SUBJECT type:} \texttt{ORG}

\textbf{OBJECT:} \textarabic{قلقيلية}

\textbf{OBJECT type:} \texttt{GPE}
\end{quote}

For input $x_i$, let $\mathcal{C}_i$ be the compatible relation classes,
including \norel{}. The classifier produces
$\mathbf{z}\in\mathbb{R}^{41}$. Ontology-incompatible logits are masked before
cross-entropy, so the effective training objective is
\begin{equation}
\mathcal{L}_{\mathrm{CLS}}
=
-\log
\frac{\exp z_{r^*}}
{\sum_{r'\in\mathcal{C}_i}\exp z_{r'}}.
\end{equation}
Rows whose gold relation is incompatible with the supplied entity types are
rejected, and the same compatibility mask is used for evaluation and final
prediction.

\begin{table}[t]
\centering
\small
\begin{tabular}{@{}p{0.28\columnwidth}
                p{0.30\columnwidth}
                p{0.30\columnwidth}@{}}
\toprule
\textbf{Property} & \textbf{Causal LM} & \textbf{Classifier} \\
\midrule
Output & Generated/scored label & One score per class \\
Objective & Token prediction & Multiclass CE \\
Inference & Generation/scoring & Masked argmax \\
Parsing & Required & Not required \\
Ontology & Prompt/candidate set & Direct logit mask \\
\bottomrule
\end{tabular}
\caption{Qwen-RE causal-LM and classifier formulations.}
\label{tab:qwen-formulations}
\end{table}

\subsection{Class-Imbalance Interventions}

For weighted causal-LM training, raw inverse-frequency weights are
\begin{equation}
w_r^{\mathrm{raw}}=\frac{1}{n_r}.
\end{equation}
They are normalized to mean one and clipped:
\begin{equation}
w_r
=
\min\left(
5.0,
\max\left(
0.25,
\frac{w_r^{\mathrm{raw}}}
{\frac{1}{|\mathcal{R}|}\sum_{r'}w_{r'}^{\mathrm{raw}}}
\right)
\right).
\end{equation}

Qwen-RE-9B-Cls instead uses a weighted random sampler with
\[
q_r\propto(1/n_r)^{0.5}.
\]

\subsection{QLoRA Fine-Tuning}

Qwen3.5-4B and Qwen3.5-9B are adapted with QLoRA
\citep{dettmers2023qlora}. The frozen backbone is loaded in 4-bit NF4 with
double quantization and BF16 computation. LoRA adapters target
\texttt{q\_proj}, \texttt{k\_proj}, \texttt{v\_proj}, \texttt{o\_proj},
\texttt{gate\_proj}, \texttt{up\_proj}, and \texttt{down\_proj}.

\begin{table}[t]
\centering
\small
\begin{tabular}{ll}
\toprule
\textbf{Hyperparameter} & \textbf{Value} \\
\midrule
Training epochs & 8 \\
Learning rate & $2\times10^{-4}$ \\
Scheduler / warm-up & Cosine / 0.03 \\
Weight decay & 0.0 \\
LoRA $r/\alpha$/dropout & 32 / 64 / 0.05 \\
Quantization & 4-bit NF4 \\
Double quantization & Yes \\
Compute dtype & bfloat16 \\
Optimizer & Paged AdamW 8-bit \\
Gradient checkpointing & Yes \\
Evaluation interval & 1,000 steps \\
Seed & 42 \\
\bottomrule
\end{tabular}
\caption{Core Qwen-RE QLoRA settings recorded by the released configurations.}
\label{tab:qwen-qlora}
\end{table}

The sequence-classification run uses train/evaluation batch sizes 4/4,
gradient accumulation 4, and maximum sequence length 1,280; the recorded
causal-LM evaluation configuration uses batch size 1. The classifier's recorded
LoRA rank and scaling are $r=32$ and $\alpha=64$.

\subsection{Validation and Checkpoint Selection}

Causal-LM checkpoints are evaluated on the 1,687 development examples using
generation-based inference. Outputs are mapped to the relation inventory, with
ontology-incompatible predictions mapped to \norel{}, and checkpoints are
selected by development micro-$F_1$.

The best confirmed development micro-$F_1$ is 0.6846 for Qwen-RE-4B and
0.6977 for Qwen-RE-9B. Table~\ref{tab:qwen-validation-app} shows the archived
trajectory for the inspected 9B run. Since this is single-label classification,
accuracy equals all-class micro-$F_1$; positive-only $F_1$ is reported
separately.

\begin{table}[t]
\centering
\small
\setlength{\tabcolsep}{3pt}
\begin{tabular}{@{}lrr@{}}
\toprule
\textbf{Checkpoint} & \textbf{Positive $F_1$} &
\textbf{Acc./micro-$F_1$} \\
\midrule
Base prompt & 0.5641 & 0.4707 \\
Step 1,000 & 0.5035 & 0.6129 \\
Step 2,000 & \textbf{0.7155} & \textbf{0.6977} \\
Step 2,943 & 0.7099 & 0.6941 \\
\bottomrule
\end{tabular}
\caption{Archived Qwen-RE-9B development results.}
\label{tab:qwen-validation-app}
\end{table}

Step 2,000 correctly predicts 1,177 of 1,687 examples. Because the archive
does not reliably map checkpoint steps to epochs, we report steps directly.
These development results are not directly comparable with the official blind
scores because they use a different split and evaluation pipeline.
\subsection{Extended Results}
\label{app:qwen-results}

\begin{table}[t]
\centering
\small
\begin{tabular}{lr}
\toprule
\textbf{Qwen-RE system} & \textbf{Official $F_1$} \\
\midrule
Qwen-RE-4B & 91.53 \\
Qwen-RE-9B & 91.76 \\
Qwen-RE-9B-Cls & 91.94 \\
\bottomrule
\end{tabular}
\caption{Official Qwen-RE results.}
\label{tab:qwen-official-app}
\end{table}

The official scores differ by 0.18 points between Qwen-RE-9B and Qwen-RE-9B-Cls.

\section{Qwen Exploratory Prompt Optimization}
\label{app:prompt-optimization}

\subsection{Disambiguation Rules}

We manually wrote decision rules for semantically close relations using the
definitions in WojoodOntology
\citep{aljabari-etal-2025-wojoodontology}. A representative fragment is:

\begin{quote}
\small
\textbf{Allowed labels and Arabic meanings:}

\rel{Location.located\_in}: ordinary physical or geographic containment.

\rel{Location.headquartered\_in}: OBJECT is the headquarters or principal
institutional location of SUBJECT.

\rel{PartOf.geopolitical\_division}: SUBJECT is an administrative or
geographic division of OBJECT.

\textbf{Decision rule:} Prefer the more specific headquarters,
administrative-division, capital, border, or proximity relation when its
evidence is present; otherwise use ordinary containment.
\end{quote}

These rules were diagnostically useful but did not eliminate confusion. In
particular, headquarters examples could still be predicted as
\rel{Location.located\_in}, while longer supportive and contrastive rules
reduced validation performance for the selected Qwen configuration.

\subsection{GEPA Prompt Evolution}

We explored GEPA, a reflective prompt-evolution method that revises prompts
using model errors and validation performance
\citep{agrawal2026gepa}. The experiment used a stratified subset of 506
training examples and 196 validation examples.

For each candidate prompt, prediction errors were summarized with the gold and
predicted labels, model output, compatible labels, relation meanings, and
disambiguation guidance. The reflective editor
\texttt{openai/gpt-5.5} used this feedback to propose revised prompts, which
were evaluated on the validation subset and selected using GEPA's Pareto-based
procedure.

The best stored GEPA prompt obtained positive $F_1=0.6011$ and accuracy
$0.5408$ on the 196-example validation subset. Because the corresponding
pre-GEPA score on the same subset was not recorded, we do not claim a measured
improvement over the initial prompt. This experiment was exploratory and did
not replace the official system.

\section{Additional Error Analysis}
\label{app:error-analysis}

\subsection{Entity-Type Errors}

A hard type error can simultaneously alter the typed text markers, the
42-dimensional subject--object type vector, and the ontology-compatible
relation inventory. Type prediction is therefore consequential even though
TypePredictor achieves high aggregate accuracy, particularly for lower-support
types and unseen-sentence mentions.

\subsection{Repeated Mentions}

When a surface form occurs multiple times, the system must associate the
provided subject and object strings with the textual occurrences that express
the relation. Nearest-candidate selection can choose an incorrect occurrence,
while score-all inference increases computation and introduces competing span
hypotheses. Joint occurrence and relation scoring is a natural direction for
future work.

\subsection{Semantically Close Relations}

A recurring difficulty is distinguishing generic and specific geographic
relations, particularly \rel{Location.located\_in},
\rel{Location.headquartered\_in}, and
\rel{PartOf.geopolitical\_division}. The much larger training support for
\rel{Location.located\_in} can favor the generic relation when the sentence
contains evidence for a more specific alternative.

\subsection{Causal Generation Errors}

Causal generation adds a separate failure mode: a model can assign useful
semantic probability mass to the correct relation while still generating
extra prose, punctuation, an alias, or only part of a canonical label. This
motivated complete-label scoring in Qwen-RE and the direct 41-way
classification formulation in Qwen-RE-9B-Cls.

\section{Reproducibility}
\label{app:reproducibility}

\subsection{Released Resources}

The current public inventory includes:

\begin{itemize}[leftmargin=*]
\item \texttt{U4RASD/TypePredictor};
\item \texttt{U4RASD/DRU-RE-TypedPair-TwoHead};
\item TypedPair checkpoints for AraModernBERT, AraBERTv02, ARBERTv2,
AraBERTv02-Large, and NeoAraBERT-no-interactions;
\item \texttt{U4RASD/DRU-RE-Yehia} and
\texttt{U4RASD/DRU-RE-Yehia-2};
\item \texttt{U4RASD/WojoodRelationsAnnotated};
\item \texttt{U4RASD/Yehia-RE-SFT};
\item \texttt{U4RASD/TypePrediction}; and
\item \texttt{U4RASD/DRU-RE-Qwen}.
\end{itemize}

The repository identifiers retain their released names even where the paper
uses the consolidated display names Yehia-RE and Qwen-RE.

\subsection{Training and Selection Summary}

\begin{table*}[t]
\centering
\small
\setlength{\tabcolsep}{4pt}

\begin{tabularx}{\textwidth}{@{}l l X X X@{}}
\toprule
\textbf{Component} &
\textbf{Backbone} &
\textbf{Training} &
\textbf{Output} &
\textbf{Selection rule} \\
\midrule

TypePredictor &
NeoAraBERT &
Full fine-tuning, 4 epochs &
21 type logits &
Validation macro-$F_1$ \\

TypedPair &
Arabic encoders &
Full fine-tuning, 4 epochs &
Existence + 40 relation logits &
Development positive micro-$F_1$ \\

Yehia-RE &
Yehia-7B &
QLoRA, 3 epochs &
One Arabic code token &
Development positive micro-$F_1$ \\

Qwen-RE-4B/9B &
Qwen3.5-4B/9B &
QLoRA, 8 epochs &
Canonical label text &
Development micro-$F_1$ \\

Qwen-RE-9B-Cls &
Qwen3.5-9B &
QLoRA + 41-way head &
Masked class logits &
Development micro-$F_1$ \\

\bottomrule
\end{tabularx}

\caption{Cross-system summary of the principal training and selection choices.
Detailed configurations are reported in the corresponding system appendices.}
\label{tab:main-config}
\end{table*}
\subsection{Software, Hardware, and Model Revisions}

The released artifacts do not contain one environment manifest shared by all
experiments. Table~\ref{tab:environment} therefore reports exact recorded
values where available and explicitly marks information that was not preserved
rather than estimating it.

\begin{table*}[t]
\centering
\small
\begin{tabularx}{\textwidth}{@{}lllX@{}}
\toprule
\textbf{System} & \textbf{Component} & \textbf{Recorded value} &
\textbf{Notes} \\
\midrule
TypePredictor & Python & 3.12.3 & Recorded environment \\
TypePredictor & PyTorch & 2.5.1+cu124 & Recorded environment \\
TypePredictor & Transformers & 4.49.0 & Recorded environment \\
TypePredictor & GPU & NVIDIA A40 & 44.43 GiB reported \\
\midrule
Yehia-RE & PyTorch & 2.5.1 & Training requirements \\
Yehia-RE & Transformers & 4.57.1 & Training requirements \\
Yehia-RE & PEFT & 0.17.1 & Training requirements \\
Yehia-RE & bitsandbytes & 0.45.5 & Training requirements \\
Yehia-RE & vLLM & 0.10.2 & Blind-test runner \\
Yehia-RE & GPU & NVIDIA L40S & Selected run validated on one GPU \\
Yehia-RE & Base revision &
\texttt{b9dda471...f4ebb8} & Full hash reported in Appendix~\ref{app:yehia} \\
\midrule
Qwen-RE & PyTorch & 2.5.1 & Training environment \\

Qwen-RE & Transformers & 4.49.0 & Training environment \\

Qwen-RE & PEFT & 0.19.1 & Recorded adapter configuration; LoRA/QLoRA adapter implementation \\

Qwen-RE & bitsandbytes & 0.50.0 & 4-bit QLoRA quantization; latest PyPI release before 31 July 2026 \\

Qwen-RE & GPU & NVIDIA RTX PRO 4500 & Training hardware \\
\bottomrule
\end{tabularx}
\caption{Software, hardware, and revision information recoverable from the
released experiment artifacts.}
\label{tab:environment}
\end{table*}

TRL is not listed as a dependency of the recorded Yehia-RE training
environment. Exact Python and CUDA toolkit versions for Yehia-RE, and the
corresponding complete software/hardware environment for Qwen-RE, are not
recoverable from the archived releases and are therefore not inferred here.

\end{document}